\documentclass[11pt,letterpaper]{article}
\usepackage[top=0.85in,left=1in,footskip=0.75in,marginparwidth=2in]{geometry}

\usepackage{amsmath}
\usepackage{amssymb}
\usepackage{amsthm}
\usepackage[sort&compress,comma,super]{natbib}
\usepackage{cases}
\usepackage{siunitx}
\usepackage{multirow}
\usepackage[utf8]{inputenc}
\usepackage{subfigure}
\usepackage{nameref,hyperref}

\usepackage{microtype}
\DisableLigatures[f]{encoding = *, family = * }

\usepackage{changepage}

\usepackage[aboveskip=1pt,labelfont=bf,labelsep=period,singlelinecheck=off]{caption}

\makeatletter
\renewcommand{\@biblabel}[1]{\quad#1.}
\makeatother

\usepackage{lastpage,fancyhdr,graphicx}
\usepackage{epstopdf}

\usepackage{xcolor}

\definecolor{Gray}{gray}{.25}

\usepackage{graphicx}
\usepackage{float}
\usepackage{makecell}
\usepackage{sidecap}

\usepackage{wrapfig}
\usepackage[pscoord]{eso-pic}
\usepackage[fulladjust]{marginnote}
\reversemarginpar

\definecolor{nicered}{HTML}{C14000}
\newcommand{\qi}[1]{\textcolor{nicered}{#1}}
\renewcommand{\qi}{}

\usepackage[textsize=tiny,
    textcolor=red,
    linecolor=red,
    bordercolor=red,
    backgroundcolor=white,disable
]{todonotes}

\begin{document}
\vspace*{0.35in}

\begin{flushleft}
{\Large
\textbf\newline{
Toward stochastic neural computing
}
}
\newline
\\
Yang Qi\textsuperscript{1,2,3}, 
Zhichao Zhu\textsuperscript{1}, 
Yiming Wei\textsuperscript{1}, 
Lu Cao\textsuperscript{4},
Zhigang Wang\textsuperscript{4},
Jie Zhang\textsuperscript{1,2},
Wenlian Lu\textsuperscript{1,2},
Jianfeng Feng\textsuperscript{1,2,*}
\\
\bigskip
\it{1} Institute of Science and Technology for Brain-Inspired 
Intelligence, Fudan University, Shanghai 200433, China
\\
\it{2} Key Laboratory of Computational Neuroscience and Brain-Inspired 
Intelligence (Fudan University), Ministry of Education, China
\\
\it{3} MOE Frontiers Center for Brain Science, Fudan University, Shanghai 200433, China
\\
\it{4} Intel Labs China, Beijing, 100190, China

\bigskip 
\end{flushleft}

\section*{Abstract} %
The highly irregular spiking activity of cortical neurons and behavioral variability suggest that the brain could operate in a fundamentally probabilistic way. Mimicking how the brain implements and learns probabilistic computation could be a key to developing machine intelligence that can think more like humans. In this work, we propose a theory of stochastic neural computing (SNC) in which streams of noisy inputs are transformed and processed through populations of nonlinearly coupled spiking neurons. To account for the propagation of correlated neural variability, we derive from first principles a moment embedding for spiking neural network (SNN). This leads to a new class of deep learning model called the moment neural network (MNN) which naturally generalizes rate-based neural networks to second order. As the MNN faithfully captures the stationary statistics of spiking neural activity, it can serve as a powerful proxy for training SNN with zero free parameters. Through joint manipulation of mean firing rate and noise correlations in a task-driven way, the model is able to learn inference tasks while simultaneously minimizing prediction uncertainty, resulting in enhanced inference speed. We further demonstrate the application of our method to Intel's Loihi neuromorphic hardware. The proposed theory of SNC may open up new opportunities for developing machine intelligence capable of computing uncertainty and for designing unconventional computing architectures. %

\section{Introduction}
The ability to represent and to compute with uncertainty is a key aspect of intelligent systems including the human brain. In classic analog computing, noise is often harmful as information carried within the signal gradually degrades as noise is introduced with each step of computation. Digital computing resolves the issue of computing with noisy amplitude by representing signal with discrete binary codes, ensuring both accuracy and resilience against errors. However, neurons in the brain communicate through highly irregular spiking activity~\citep{TOMKO1974405,TOLHURST1983775,softky1993highly} which is noisy in time domain but not in its amplitude. Furthermore, neural and behavioral responses also exhibit trial-to-trial variability even with identical stimulus~\citep{arieli1996dynamics}. It is one of nature's great mysteries how such a noisy computing system like the brain can perform computation reliably. Mimicking how the brain handles uncertainty may be crucial for developing intelligent agents~\citep{Gerven2017frontiers} and more efficient computing systems~\cite{Maass2014ieee}. 

One of the prominent ideas is that neural computing is inherently stochastic and that noise is an integral part of the computational process in the brain rather than an undesirable side effect~\citep{DECO20091,Fiser2010,Maass2014ieee}. Stochastic neural dynamics is implicated in a broad range of brain functions from sensory processing~\citep{knill1996,YUILLE2006301}, cognitive tasks~\citep{griffiths2007topics,vul2014one} to sensorimotor control~\citep{wolpert2007probabilistic,Kording2004nature}, and is theorized to play important roles in computational processes such as uncertainty representation~\citep{Ma2014a,henaff2020representation}, probabilistic inference~\citep{Deneve2008neco,Ma2006,Hoyer2002nips}, and neural population coding~\citep{panzeri1999correlations,Kohn2016,ding2011dynamic,averbeck2006neural}.

\qi{A fundamental question is how the brain might implement probabilistic computation at the neural network level~\citep{Lin2022bayes_inf_review,Pouget2013a}. To address this, a diversity of theories including probabilistic neural population coding~\citep{Beck2011,Ma2006} and neural sampling theories~\citep{Hoyer2002nips,Buesing2011ploscb,Orban2016,Qi2022nat_comm,Aitchison2016hamiltonian} have been proposed. Of particular interest is how probabilistic computation might be implemented using the inherently irregular dynamics of spiking neural activity~\citep{Deneve2008neco,Qi2022nat_comm}. A popular choice is considering binary or Poisson-like neuron models in which the probability of firing for each neuron is specified in terms of its inputs~\citep{Buesing2011ploscb,Jang2019}. However, by making this simplification, these models essentially erase the correlation structure that may naturally arise from the synaptic coupling. This is problematic because mean firing rate is nonlinearly coupled to noise correlation in biological neurons~\citep{Rocha2007nature} and neural coding theories also suggest that correlated neural activity may significantly impact neural population coding~\citep{schneidman2006weak,Kohn2016,Panzeri2022,averbeck2006neural,ma2022dynamics,Valente2021nat_neurosci}. The main challenge lies in representing and manipulating high-dimensional joint probability distributions of a large population of spiking neurons with realistic firing mechanism.}

\qi{A point of contention regards the origin of irregular or fluctuating neural activity for modeling probabilistic neural computation. One school of thought suggests that these irregularity could be `explained away' by intrinsic noise of neurons, such as stochasticity in neural firing or synaptic noise. This is the traditional view taken by modeling studies on Bayesian neural networks~\citep{Jospin2022IEEE} and on probabilistic spiking neural networks with Poisson-like neurons~\citep{Buesing2011ploscb,Jang2019,Ma2023patterns}. It has also been suggested that irregular spiking activity could reflect the random motion of spatio-temporal activity patterns emerging from the recurrent dynamics of a neural circuit~\citep{Keane2015jneursci,Qi2022nat_comm}. Alternatively, irregular neural activity could reflect uncertainty in sensory inputs whereas the neural network model itself is deterministic~\citep{Deneve2008neco}. It remains unclear how a network of spiking neurons with deterministic dynamics is able to process probabilistic computation over a temporal stream of noisy data.}

\qi{While theoretical modeling of probabilistic neural computation often rely on hand-crafted models with varying degrees of abstraction~\citep{Lin2022bayes_inf_review,Pouget2013a}, there is a growing interest in how general probabilistic neural computation can be learned~\cite{Ma2023patterns,Jang2019,Dutta2022nat_comm,Orhan2017,Ichikawa2022neco,Quax2021sci_rep_prior}. This problem is traditionally considered in the setting of Bayesian neural networks, which can be trained using Markov chain Monte Carlo sampling or variational inference~\citep{Dutta2022nat_comm}, but it has also been demonstrated that probabilistic tasks could be learned by deterministic, rate-based neural network models~\citep{Quax2021sci_rep_prior,Orhan2017,Ichikawa2022neco}. 
For spiking neural networks, one line of research focuses on spike-time dependent plasticity (STDP) using simplified probabilistic spiking neuron models~\citep{Nessler2013plos_cb,Rezende2014frontiers}. While STDP is more biological plausible, it is generally not competitive to backpropagation in performance. More recently, learning in deterministic spiking neural networks with integrate-and-fire dynamics has been intensively studied~\citep{Pfeiffer2018_deep_learning,Neftci2019IEEE}, leading to a range of learning algorithms including direct training methods and ANN-to-SNN conversion. Direct training methods typically employ some variants of backpropagation-through-time over the membrane potential dynamics which could be expensive to train~\citep{BOHTE2002neurocomputing,Lee2016,10.3389/fnins.2018.00331,shrestha2018slayer,Bellec2020,rathi2020enabling,10.3389/fnins.2020.00119,Yan2021,Wunderlich2021sci_rep,Yan2022}. While ANN-to-SNN conversion is more scalable, this is often achieved by creating \textit{ad hoc} approximate mappings between ANN and SNN or through post-training optimization~\citep{Diehl2015,10.3389/fnins.2017.00682,Hunsberger2015arxiv,yan2021near,ding2021optimal,Kim22811}, but so far a general guiding principle is lacking. Moreover, these works typically assume firing rate coding or temporal coding, and learning in spiking neural networks under probabilistic coding has been largely unexplored. %
}

\qi{In this work, we develop a theory of stochastic neural computing (SNC) based on spiking neuron models with biologically realistic firing mechanism and show how high-dimensional joint probability distributions can be represented and transformed through nonlinearly coupled, correlated spiking activity of large populations of neurons. We overcome the aforementioned challenges for modeling and learning SNC by developing a novel moment embedding approach for spiking neurons accounting for correlated neural variability.} In this approach, spike trains are first mapped to their statistical moments up to second order, to obtain a minimalistic yet accurate statistical description of neural spike train~\citep{Feng2006,LU2010913}. Unlike in standard firing rate models, signal and noise are concurrently processed through neural spikes. \qi{This leads to a new class of deep learning model called the moment neural network (MNN) which naturally generalizes rate-based artificial neural networks to second order.} We implement gradient-based learning in this model and incorporate uncertainty into the learning objective, thus enabling direct manipulations of correlated neural variability in a task-driven way. The synaptic weights obtained this way can be used directly, without further fine tuning of free parameters, to recover the original spiking neural network.

To demonstrate, we train a spiking neural network to perform an image classification task using a feedforward architecture. Through minimizing a generalized cross entropy, the model is able to learn the task while simultaneously minimizing trial-to-trial variability of model predictions. The trained network naturally exhibits realistic properties of cortical neurons including mean-dominant and fluctuation-dominant activities as well as weak pairwise correlations. We reveal concurrent and distributed processing of signal and noise in the network and explain how structured neural fluctuations lead to \qi{both accurate inference and enhanced speed}. We further demonstrate applications of the proposed method on neuromorphic hardware and explain how SNC may serve as a guiding principle for future design of neuromorphic computing.
 \section{Results}
\subsection{A probabilistic interpretation of spike-based neural computation}

\begin{figure}
\centering
\includegraphics[width=0.9\textwidth]{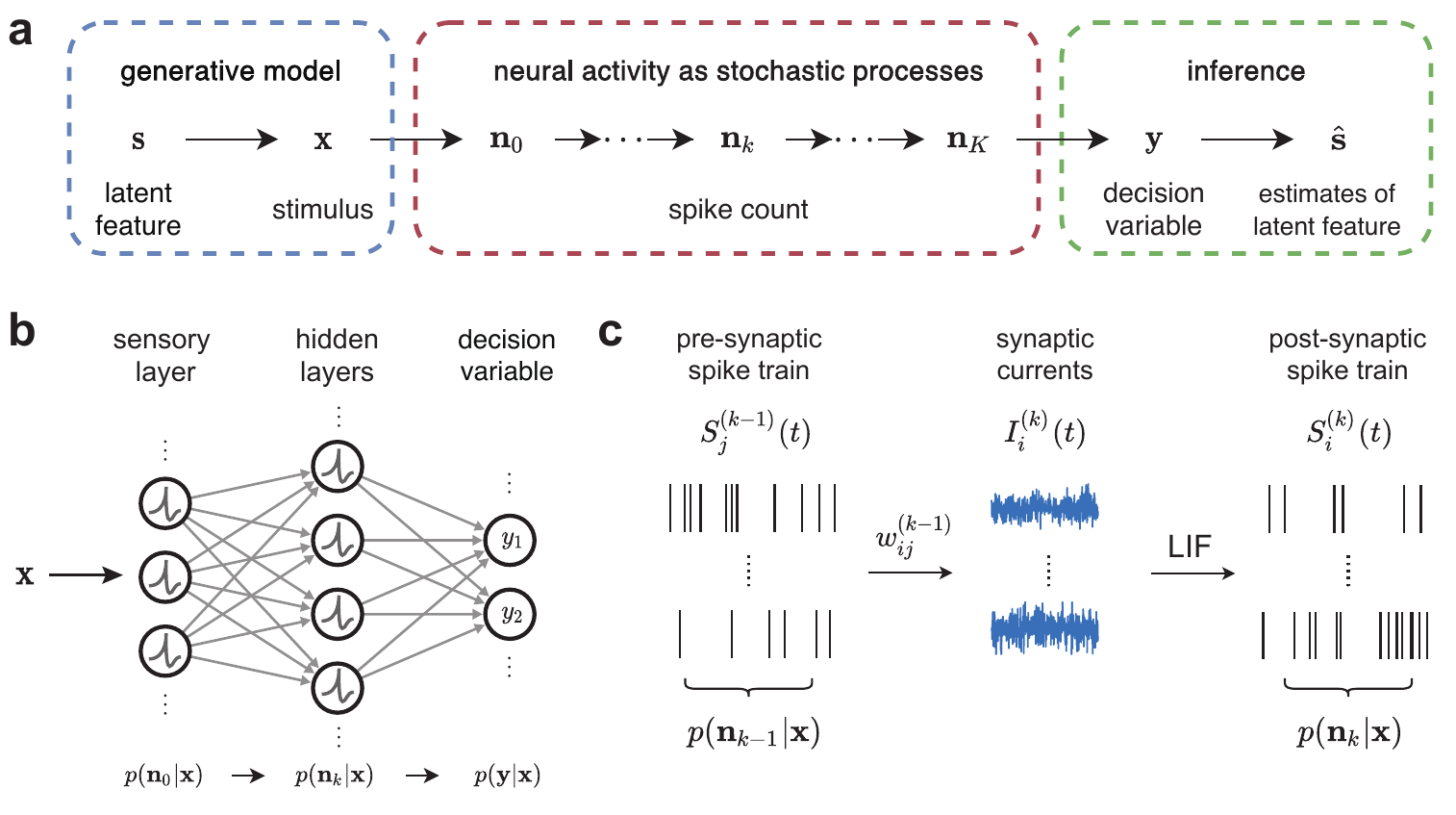}\\
\caption{\textbf{Spike-based stochastic neural computing (SNC).} 
\textbf{a}, A graphical model for SNC which consists of a generative model describing the external environment, multiple stages of fluctuating neural activity states, and a readout for making inference. The index $k$ represents different neural populations in a feedforward network or alternatively time in a recurrent network. 
\textbf{b}, Schematics of a spiking neural network implementing the computational processes outlined in \textbf{a}, with each layer characterized by a joint probability distribution of neural spike count. 
\textbf{c}, Propagation of irregular neural spike trains through two feedforwardly connected neural populations. The pre-synaptic spike trains first undergo synaptic summation to generate fluctuating synaptic currents, which in turn drive post-synaptic neurons to fire. The probability distribution of spike count is transformed in a non-trivial way due to the nonlinear coupling of correlated neural variability.
\label{fig:mnn_schematic} }
\end{figure}

\qi{
To establish an understanding about the computational goal of our stochastic neural computing (SNC) theory, consider the probabilistic generative model shown in Fig.~\ref{fig:mnn_schematic}\textbf{a}, in which a stimulus $\mathbf{x}$ (such as an image) is characterized by a latent feature $\mathbf{s}$ (such as the category the image belongs to). Conceptually, this generative model can be represented by a conditional probability distribution $p(\mathbf{x}|\mathbf{s})$, that is, the likelihood function for $\mathbf{s}$. The goal is then to infer $\mathbf{s}$ given any stimulus $\mathbf{x}$ as the input, or more precisely, a posterior distribution $p(\mathbf{s}|\mathbf{x})$ of $\mathbf{s}$ conditioned on $\mathbf{x}$~\citep{Orban2016}. For optimal Bayesian inference, the posterior obeys Bayes' rule
\begin{equation}
p(\mathbf{s}|\mathbf{x})
=\dfrac{p(\mathbf{x}|\mathbf{s})p(\mathbf{s})}{p(\mathbf{x})},
\label{eq:bayes_rule}
\end{equation}
where $p(\mathbf{s})$ represents the prior distribution.  
}

\qi{
The observations that neurons in the brain exhibit irregular spiking activity and that behavioral response exhibits trial-to-trial variability have led to the Bayesian brain hypothesis~\citep{knill1996}, suggesting that the brain may perform probabilistic inference. Whether human performs probabilistic inference in a Bayes-optimal way is a question of debate~\citep{Adler2018plos_cb,Li2020nat_comms}, therefore we will adopt the broader notion of probabilistic inference without requiring Bayesian optimality. In practice, the probabilistic generative model about the environment is unknown, and what we often want is to quantify and minimize prediction uncertainty. Nonetheless, equation~\ref{eq:bayes_rule} provides a useful conceptual guide for the computational goal of SNC. The key idea is that the brain computes not only point estimates about latent features but also uncertainty or confidence about it.
}

\qi{
A central question under the Bayesian brain hypothesis is how the brain implements probabilistic inference. One of the prominent theories is probabilistic neural population code, in which certain probability operations such as integrating information from multiple sensory channels can be mapped to the corresponding operations with neural activity~\citep{Dehaene2021plos_cb,Ma2006,walker2020neural}. Another class of theories are based on the neural sampling hypothesis suggesting that temporally fluctuating neural activity implements probabilistic sampling over time~\cite{Orban2016,Qi2022nat_comm}. A number of studies have considered probabilistic inference with a special class of probabilistic generative model, the Gaussian scale mixture (GSM) model, whose posterior distribution has known analytical solutions. It has been shown analytically that the GSM model can be mapped to a linear stochastic neural network model for implementing Hamiltonian Monte Carlo sampling~\citep{Aitchison2016hamiltonian}. In a different study, a nonlinear stochastic rate-based neural network is trained to perform inference under the GSM model, albeit the neural network considered has a specific connectivity structure with a small number of parameters~\citep{echeveste2020cortical}. It has also been suggested that neural sampling may be implemented using spatiotemporal activity patterns of spiking neurons~\citep{Qi2022nat_comm}. Despite a plethora of theoretical proposals, how arbitrary probabilistic inference tasks can be implemented at the neural network level remains elusive. 
}

\qi{
Recent advances in deep learning has opened up new opportunities for implementing probabilistic inference in neural systems through learning~\citep{Soo2022nips}. In a previous study, rate-based artificial neural network models (ANNs) are trained to perform probabilistic tasks~\citep{Orhan2017}. An alternative approach maps mean firing rates to the sufficient statistics of an exponential family distribution~\citep{Vertes2018nips_ddc}. The issue with using ANNs is that they generally lack a native representation of uncertainty through the intrinsic fluctuations as seen in biological neurons. An alternative approach considers stochastic binary spiking neural network with firing probability depending on the neuron's input~\cite{Ma2023patterns,Jang2019}. However, this approach typically involves simplified binary neurons with uncorrelated spiking activity and it is unclear how more general SNC with correlated neural activity can be implemented in biologically plausible spiking neural network models.
}

Given that the computational goal of SNC is using neural networks to perform probabilistic inference, we now introduce its general theoretical foundations. Consider a computational process shown in Fig.~\ref{fig:mnn_schematic}\textbf{a}, which consists of three components. The first component is a generative model describing how an observable stimulus $\mathbf{x}$ in the environment depends on its latent features $\mathbf{s}$. The second component is a model describing the fluctuating activity states $\mathbf{n}_{k}$ of a group of neural populations, which are interpreted as random spike counts over a time window $\Delta t$. The index $k$ represents different neural populations in a feedforward network or alternatively discrete time steps in a recurrent network. The last component is a decision variable or readout $\mathbf{y}$ \qi{used for producing estimates $\hat{\mathbf{s}}$ about the latent features $\mathbf{s}$.}

To express these computation stages concretely, we write down the distribution $p(\mathbf{n}_k\vert \mathbf{x})$ of the neural population state $\mathbf{n}_k$ at each stage $k$ in terms of the marginalization of its conditional probability over $\mathbf{n}_{k-1}$ in the preceding stage as
\begin{equation}
p(\mathbf{n}_k\vert \mathbf{x})=\int p(\mathbf{n}_k\vert\mathbf{n}_{k-1})p(\mathbf{n}_{k-1}\vert \mathbf{x})d\mathbf{n}_{k-1}.
\label{eq:one_step}
\end{equation}
\qi{Note that the spike count time window is assumed to be sufficiently large relative to the temporal correlation time scale so that $p(\mathbf{n}_k\vert\mathbf{n}_{k-1})$ does not involve spike count at earlier times}. By chaining equation (\ref{eq:one_step}) iteratively, we recover the probability of the readout $\mathbf{y}$ conditioned on the stimulus $\mathbf{x}$
\begin{equation}
p(\mathbf{y}\vert \mathbf{x})
= \int p(\mathbf{y}\vert \mathbf{n}_K)
\prod_{k=1}^{K} p(\mathbf{n}_{k}\vert \mathbf{n}_{k-1})
p(\mathbf{n}_0\vert \mathbf{x}) d\mathbf{n}_{0}\cdots d\mathbf{n}_{K}.
\label{eq:markov_chain}
\end{equation}
Equipped with this conceptual framework, we can now define SNC as a series of neural operations [equation (\ref{eq:one_step})] that generates a desired conditional distribution $p(\mathbf{y}\vert \mathbf{x})$ of the readout $\mathbf{y}$ given a stimulus $\mathbf{x}$. Under this view, the fundamental computing unit of SNC is the probability distribution of the activity state of a neural population $p(\mathbf{n})$, and the basic operation of SNC is the transformation of these distributions across populations of neurons. \qi{
In general, the transition probability $p(\mathbf{n}_{k}\vert \mathbf{n}_{k-1})$ depends on the model parameters (synaptic weights), which can be optimized through learning such that the model can generate correct estimates $\hat{\mathbf{s}}$. In previous studies of probabilistic neural computation, the transition probability $p(\mathbf{n}_{k}\vert \mathbf{n}_{k-1})$ is often explicitly defined as independent Poisson or binary distributions with a firing probability depending on the input~\citep{Buesing2011ploscb,Jang2019}. However, by making this simplification, these models essentially erase the correlation structures that naturally arise from synaptic coupling. For spiking neural networks with realistic firing dynamics, the high-dimensional joint transition probability $p(\mathbf{n}_k\vert\mathbf{n}_{k-1})$ generally does not have simple expressions.} %

A spiking neural network implementing this computational process is illustrated in Fig.~\ref{fig:mnn_schematic}\textbf{b} where each neuron in the network is modeled as a leaky integrate-and-fire (LIF) neuron [see equation (\ref{eq:LIF})-(\ref{eq:current}) in Methods]. One step of SNC carried out across two populations of spiking neurons is shown in more detail in Fig.~\ref{fig:mnn_schematic}\textbf{c}. As irregular spike trains from the pre-synaptic population converge at post-synaptic neurons, they give rise to fluctuating synaptic currents and subsequently irregular spike emissions in the post-synaptic neurons~\citep{Rocha2007nature}. Importantly, since these fluctuating synaptic currents are generated from a common pool of pre-synaptic neurons, they inevitably become correlated even if the input spikes are not. These correlated neural fluctuations are further transformed in a nonlinear fashion as they propagate across downstream neural populations. %

To perform any useful computation, the spiking neural network needs to learn the set of parameter values $\theta$ that matches the readout distribution 
$p(\mathbf{y}\vert \mathbf{x};\theta)$ 
with a desired target distribution $p^*(\mathbf{y}\vert \mathbf{x})$. The probabilistic interpretation of the readout $\mathbf{y}$ allows us to design learning objectives (loss functions) in a principled manner~\citep{Jang2019}. Here we prescribe two such loss functions for regression and classification tasks under the supervised learning setting. For regression problems, a natural choice is the negative log likelihood
\begin{equation}
L(\theta)=-\sum_{\mathbf{x}\in\mathcal{D}}\log p(\mathbf{y}^*\vert\mathbf{x};\theta),
\label{eq:general_loss1}
\end{equation}
where $p(\mathbf{y}^*\vert\mathbf{x};\theta)$ is the likelihood of the network parameters $\theta$ for when $\mathbf{y}=\mathbf{y}^*$ with $\mathbf{y}^*$ representing target output. For classification problems, class prediction is obtained by taking the class label $i$ corresponding to the largest entry of $\mathbf{y}$. The probability that the model predicts class $i$ for a given input $\mathbf{x}$ can be expressed as 
$q_i(\theta) = \int p(\mathbf{y}\vert\mathbf{x};\theta)\mathbf{1}_{D_i}(\mathbf{y})d\mathbf{y}$,
where the indicator function $\mathbf{1}_{D}(\mathbf{y})$ is equal to one if $\mathbf{y}\in D$ and zero otherwise, and $D_i$ denotes the set of all $\mathbf{y}$ whose largest entry is $y_i$. Denoting $t$ as the target class, the goal is then to maximize the probability of correct prediction $q_t$. This leads to the loss function
\begin{equation}
L(\theta) = -\sum_{\mathbf{x}\in\mathcal{D}}\log(q_t)
=-\sum_{\mathbf{x}\in\mathcal{D}}\int p(\mathbf{y}\vert\mathbf{x};\theta)\mathbf{1}_{D_t}(\mathbf{y})d\mathbf{y},%
\label{eq:general_loss2}
\end{equation}
Interestingly, the same expression can be alternatively obtained from cross entropy $H(\theta) = -\sum_i p_i\log(q_i) = -\log(q_t)$ with $p_i = 1$ for $i=t$ and zero otherwise. Our formulation can thus be considered as a natural generalization of cross entropy loss commonly used in deterministic artificial neural networks. The physical significance of equation (\ref{eq:general_loss2}) is that by minimizing $L(\theta)$ the spiking neural network can be trained to make correct predictions while simultaneously minimizing trial-to-trial variability.

Although the general theoretical framework presented above provides a useful conceptual guide, question remains as how the spiking neural network under such probabilistic representation can be computed and trained. Direct evaluation of equation (\ref{eq:markov_chain}) is computationally infeasible at large scale and it is unclear how learning algorithms such as backpropagation can be implemented with respect to fluctuating neural activity. To resolve this problem, we propose a moment embedding which parameterizes the probability distributions of neural spiking activity in terms of their first- and second-order statistical moments.
\subsection{A moment embedding for gradient-based learning in spiking neural network}

\begin{figure}[h]
\centering
\includegraphics[width=0.9\textwidth]{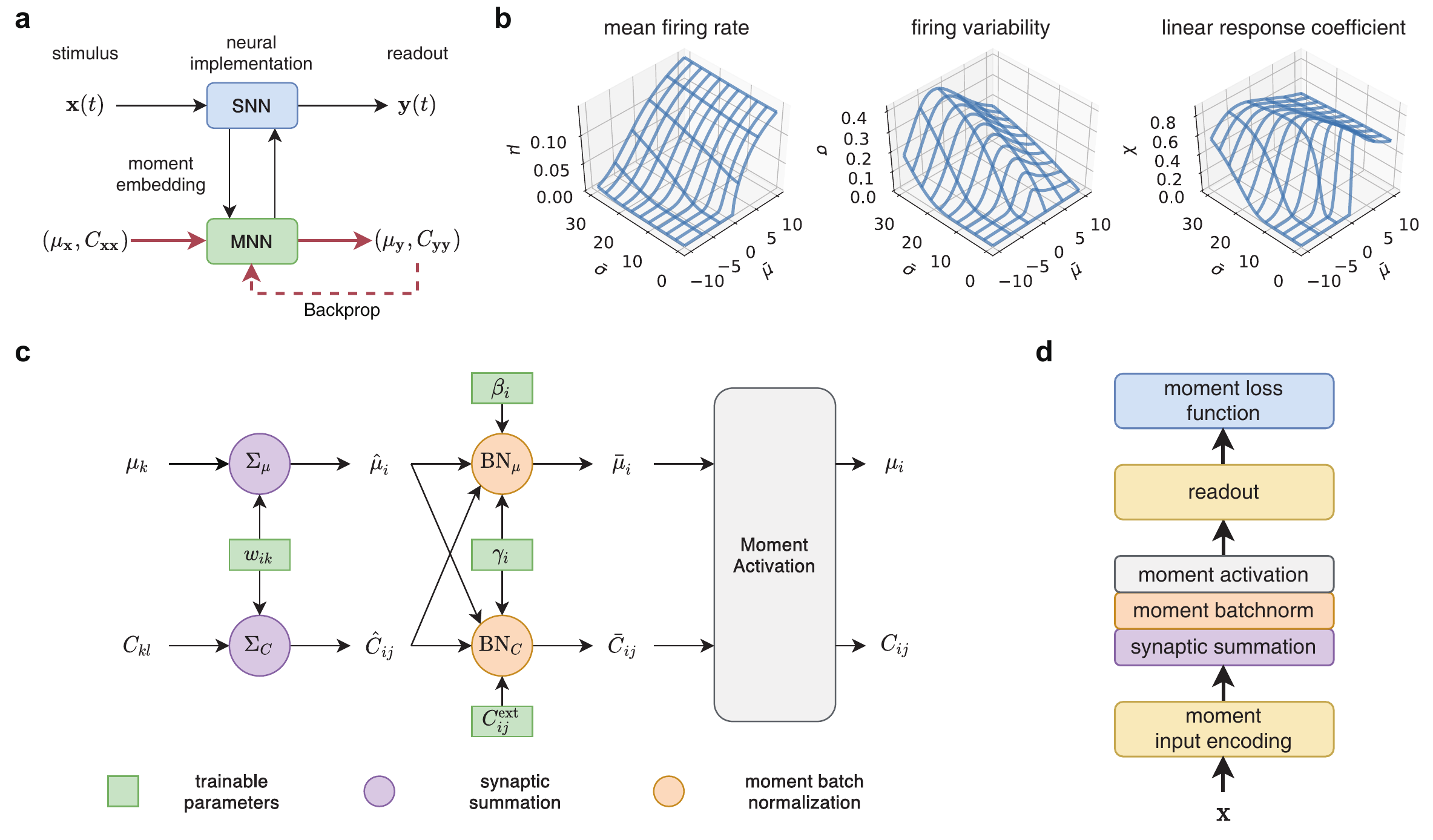}
\caption{\textbf{Gradient-based learning in spiking neural network through moment embedding}. 
\textbf{a}, Overall schematics. The spiking neural network model is first mapped to a corresponding moment neural network which can be trained with backpropagation; the trained weights are used to recover the original spiking neural network. 
\textbf{b}, Components of the moment activation function including mean firing rate $\mu$, firing variability $\sigma$, and 
the linear response coefficient $\chi$, each of which is a function of the input current mean $\bar{\mu}$ and variability $\bar{\sigma}$. In conventional analog and digital computing, such noise coupling is considered detrimental to information carried within the signal. In contrast, stochastic computing actively exploits correlated variability as a part of the computational processes. 
\textbf{c}, Computational graph of a single feedforward layer of the moment neural network, featuring synaptic summation, moment batch normalization and moment activation. 
\textbf{d}, Illustration of a trainable moment neural network with a feedforward architecture consisting of an input layer, arbitrary number of hidden layers, a readout layer, and a moment loss function. 
\label{fig:comp_graph}}
\end{figure}

The moment embedding characterizes fluctuating neural spike count with its first- and second-order statistical moments, that is, the mean firing rate $\mu$ and the firing co-variability $C$. Through a diffusion formalism~\citep{Fourcaud2002}, we can derive on a mathematically rigorous ground the mapping from the statistical moments of the pre-synaptic neural activity to that of the synaptic current, and from the synaptic current to the post-synaptic neural activity. This leads to a class of neural network models known as the \textit{moment neural network} (MNN) which faithfully captures spike count variability up to second-order statistical moments~\citep{Feng2006,LU2010913}. This can be considered as a minimalistic yet rich description of stochastic neural dynamics characterizing all pairwise neural interactions. The moment embedding essentially provides a finite-dimensional parameterization of joint probability distributions of neural spiking activity through which gradient-based learning can be performed. The network parameters trained this way can then be used directly to recover the spiking neural network without fining tuning of free parameters. An overall schematic illustrating this concept is shown in Fig.~\ref{fig:comp_graph}\textbf{a}. 

\qi{
In the following, we present the main results of the moment embedding for the leaky integrate-and-fire (LIF) spiking neuron model, though the general approach is applicable to any type of spiking neuron model, and construct the basic building blocks of the MNN, namely, synaptic summation, moment activation, moment batch normalization, and moment loss functions. The main results are summarized in Table~\ref{tab:comparison}. In the following, we explain how the moment embedding for each of these building blocks are derived, with a great emphasis placed on the correspondence between SNN, MNN, and ANN. Our analysis offers new insights to the connections between spiking and continuous-valued neural network models through the lens of stochastic neural computing. Full details of the derivation are presented in Methods [see equation (\ref{eq:def_mean}-\ref{eq:ma_chi})]. 
}

\qi{
\emph{Synaptic summation}\quad
Synaptic summation for spiking neurons involves linearly transforming the pre-synaptic spike trains $S(t)$ by the synaptic weight $W$ to obtain the post-synaptic current $I(t)$. Its moment embedding then corresponds to a mapping from the statistical moments of pre-synaptic neuron spikes to that of the post-synaptic currents. For the mean mapping, the synaptic summation works similarly as in standard rate models by calculating the synaptic current mean as a weighted sum of the pre-synaptic mean firing rate [equation (\ref{eq:sum_mean})]. Unlike rate models, however, the same synaptic weights are also used to transform the second-order moments [equation (\ref{eq:sum_cov})], resulting in correlated synaptic currents even if the input spikes are uncorrelated. 
}

\qi{
\emph{Moment activation}\quad When a spiking neuron receives fluctuating synaptic currents, it generates irregular spiking activity. There then exists a mapping from the statistical moments of the synaptic current to that of the spiking activity of post-synaptic neurons. An important feature of this moment mapping is that the mean and covariances are nonlinearly coupled, as found in cortical neurons in the brain ~\citep{Rocha2007nature}. This moment mapping, termed the moment activation, can be derived analytically from spiking neuron models via a combination of mathematical techniques including diffusion approximation and linear response theory~\citep{Feng2006,LU2010913,Rocha2007nature}. The moment activation for the leaky integrate-and-fire neuron model is given by equation (\ref{eq:ma_mu}-\ref{eq:ma_chi}). 
Figure~\ref{fig:comp_graph}\textbf{b} illustrates the three components of the moment activation, namely, the mean firing rate, the firing variability, and the linear response coefficient. An efficient numerical algorithm with custom gradients is used for implementing the moment activation~\citep{Qi2022efficient}.
}

\renewcommand{\arraystretch}{1.3}
\begin{table}[H] %
  \centering %
  \caption{Relationship between spiking neural network (SNN), rate-based artificial neural network (ANN), and moment neural network (MNN) through the lens of stochastic neural computing. The MNN systematically generalizes conventional ANN to second-order and also serves as a tight approximation to SNN. See Methods for full details.}
  \label{tab:comparison} %
  \begin{tabular}{@{} p{2.5cm}  p{4.3cm} p{3.7cm} p{5cm} @{}} %
    \hline\hline %
     & SNN & rate-based ANN & MNN \\ %
    \hline %
    variable of interest & spike count: & mean firing rate: & mean firing rate and firing covariability: \\ %
      & $\mathbf{n}(\Delta t)=\int_0^{\Delta t}S(t')dt'$ & $\mu = \displaystyle\lim_{\Delta t\to\infty}\dfrac{\mathbb{E}[\mathbf{n}]}{\Delta t}$ & 
    $\mu = \displaystyle\lim_{\Delta t\to\infty}\tfrac{\mathbb{E}[\mathbf{n}]}{\Delta t}$ \\
    & & & $C = \displaystyle\lim_{\Delta t\to\infty}\tfrac{\mathrm{Cov}[\mathbf{n},\mathbf{n}]}{\Delta t}$
    \\ %
    \hline 
    synaptic summation & linear transformation of spike train by synaptic weights: &linear transformation of mean firing rate with synaptic weights: & linear/bilinear transformation of mean firing rate/covariance by synaptic weights:\\ %
    & $I(t)=WS(t)+I_{\rm ext}$ & $\hat{\mu}=W\mu+\mu_{\rm ext}$ & $\hat{\mu}=W\mu+\mu_{\rm ext}$ \\
    & &  & $\hat{C}=WCW^T+C_{\rm ext}$\\
    \hline %
    normalization & normalizing post-synaptic current: & batchnorm: & moment batchnorm:\\ %
      & $\tfrac{I - \mathbb{E}[I]}{\sqrt{\mathrm{Var}[I]}}\gamma + \beta$  & $\bar{\mu}=\mathrm{BN}(\hat{\mu})$ & $(\bar{\mu},\bar{C})=\mathrm{MBN}(\hat{\mu},\hat{C})$\\ %
    \hline %
    activation & spiking dynamics: & pointwise activations (e.g. sigmoid, ReLU): & moment activation: \\ 
   &         $\tfrac{dV}{dt}=-LV+I,$         & $\mu=\phi(\bar{\mu})$     & $(\mu,C)=\phi(\bar{\mu},\bar{C})$ \\
       &         $V\leftarrow V_{\rm res}$ when $V>V_{\rm th}$         &      &  \\
    \hline %
   loss functions  & negative log-likelihood: & mean squared error: & moment mean squared error:\\ 
     & $-\log p(\mathbf{y}|\mathbf{x},\theta)$ & $\Vert \mu-\mathbf{y}^*\Vert ^2$ & $(\mu-\mathbf{y}^*)^T C^{-1}(\mu-
\mathbf{y}^*)\Delta t + \log \det (\tfrac{2\pi}{\Delta t} C)$ \\ 
     & probability of correct predictions: & cross-entropy: & moment cross-entropy:\\
    & $\int p(\mathbf{y}\vert\mathbf{x},\theta)\mathbf{1}_{D_i}(\mathbf{y})d\mathbf{y}$ & $-\log \sigma_t(\mu;\beta)$ & $-\log\sum_{n=1}^N \sigma_t(\tfrac{1}{\sqrt{\Delta t}}L\mathbf{z}^n+\mu;\beta)$\\ 
    \hline %
  \end{tabular}
\end{table}

\qi{
\emph{Moment batch normalization}\quad In deep networks, when the inputs are sufficiently strong (or weak), it may cause the saturation (or vanishing) of the moment activation function and subsequently the failure of gradient propagation. For conventional rate-based activation functions such as sigmoid functions, this vanishing-gradient problem is effectively alleviated through batch normalization~\citep{Ioffe2015}. Here, we propose a generalized batch normalization incorporating second-order moments, referred to as the moment batch normalization [see equation (\ref{eq:bn_mean})-(\ref{eq:bn_cov}) in Methods]. A key property of the moment batch normalization is that a common normalization factor is shared between the mean and variance of the synaptic current. Remarkably, it can be shown that the MBN is equivalent to a scaling and biasing over the post-synaptic current in the spiking neural network. As a result, the scaling and bias parameters can be re-absorbed into the synaptic weights and external input currents respectively after training is complete, thereby preserving the structure of the original spiking neural network [see equation~\ref{eq:shift_current})-(\ref{eq:scale_weight})]. 
}

\qi{
\emph{Moment loss functions}. To enable gradient-based learning, we also need to specify an appropriate loss function. The general framework of stochastic neural computing offers a principled approach to deriving loss functions incorporating second-order moments. For regression problems, we derive the moment mean squared error [MMSE, equation (\ref{eq:mmse})] based on the principle of maximum likelihood [equation (\ref{eq:general_loss1})]. This loss function simultaneously minimizes the difference between the output mean $\mu$ and the target $\mathbf{y}^*$ (systematic error) in the first term as well as the output covariance (random error) in the second term. The readout time $\Delta t$ controls the trade-off between accuracy and precision, that is, a smaller $\Delta t$ prioritizes reducing the random error more than the systematic error and vice versa. Interestingly, equation (\ref{eq:mmse}) can be interpreted as a form of free energy, such that the first and the second terms correspond to the energy and the entropy of the system, respectively. The standard mean-squared error (MSE) loss is a special case of equation (\ref{eq:mmse}) for when $C=I$. 
}

\qi{
For classification problems, a commonly used loss function in conventional ANNs is the softmax cross entropy, in which model predictions are formally assigned to a probability via the softmax function, often with analogy drawn to the Boltzmann distribution from statistical physics. However, whether the output of the softmax function can truly be interpreted as probability is questionable, since there is no underlying random processes involved. In contrast, the framework of SNC allows us to define classification accuracy in the native language of probabilities, that is, trial-wise probability of correct predictions [equation (\ref{eq:general_loss2})], whose moment embedding leads to the moment cross-entropy (MCE) loss [equation (\ref{eq:loss_CE})]. The standard cross-entropy loss used in ANNs turns out to be a special case of the MCE when the readout time is unlimited, that is, when $\Delta t\to\infty$. Remarkably, our analysis reveals the true nature of the softmax function as an approximation to the indicator function appearing in equation~(\ref{eq:general_loss2}), rather than representing some probability distribution as commonly misinterpreted. %
}

Given these basic building blocks of MNN, we can now assemble a single feedforward layer of MNN by connecting synaptic summation, moment batchnorm, and moment activation, as illustrated in Fig.~\ref{fig:comp_graph}\textbf{c}. By mapping the mean firing rate and firing covariability of the pre-synaptic population to that of the post-synaptic population, this feedforward layer essentially implements one step of stochastic neural computing in equation (\ref{eq:one_step}) which maps $p(\mathbf{n}_{k-1}\vert \mathbf{x})$ to $p(\mathbf{n}_k\vert \mathbf{x})$. Multiple hidden layers can be stacked together to form a network of arbitrary depth, effectively implementing the chain of probabilistic computation in equation (\ref{eq:markov_chain}). To enable end-to-end learning, it is also necessary to specify an appropriate moment embedding for the input and the readout for making inference. Here we assume independent Poisson input encoding, though our approach allows for any covariance structures within the input. For inference, we apply a linear readout on the last hidden layer [equation (\ref{eq:mnn_readout_mean})-(\ref{eq:mnn_readout_cov})], which is similar to synaptic summation. Figure ~\ref{fig:comp_graph}\textbf{d} shows an example of a complete feedforward MNN consisting of a Poisson-encoded input layer, a hidden layer, a linear readout, and a moment loss function. 

In summary, the moment embedding presented above leads to a new class of neural network model, the moment neural network, which is derived from SNN on a mathematically rigorous ground and at the same time naturally generalizes ANNs to second-order. The MNN serves as a conceptual link between ANN and SNN which has been missing in the literature, and provides a unifying perspective of the relationship between spiking and continuous-valued neural network models through the lens of stochastic neural computing.

\subsection{Stochastic neural computing with correlated variability}

Having developed the basic building blocks of MNN, we now demonstrate our learning framework for SNC with a classification task. For illustrative purposes, we consider a fully connected, feedforward MNN for implementing supervised learning on the MNIST datasest~\citep{deng2012mnist} consisting of images of hand-written digits. A single hidden layer and a Poisson-rate input encoding scheme are used. See Methods for details of model set-up.

\begin{figure}[H]
\centering
\includegraphics[width=0.85\textwidth]{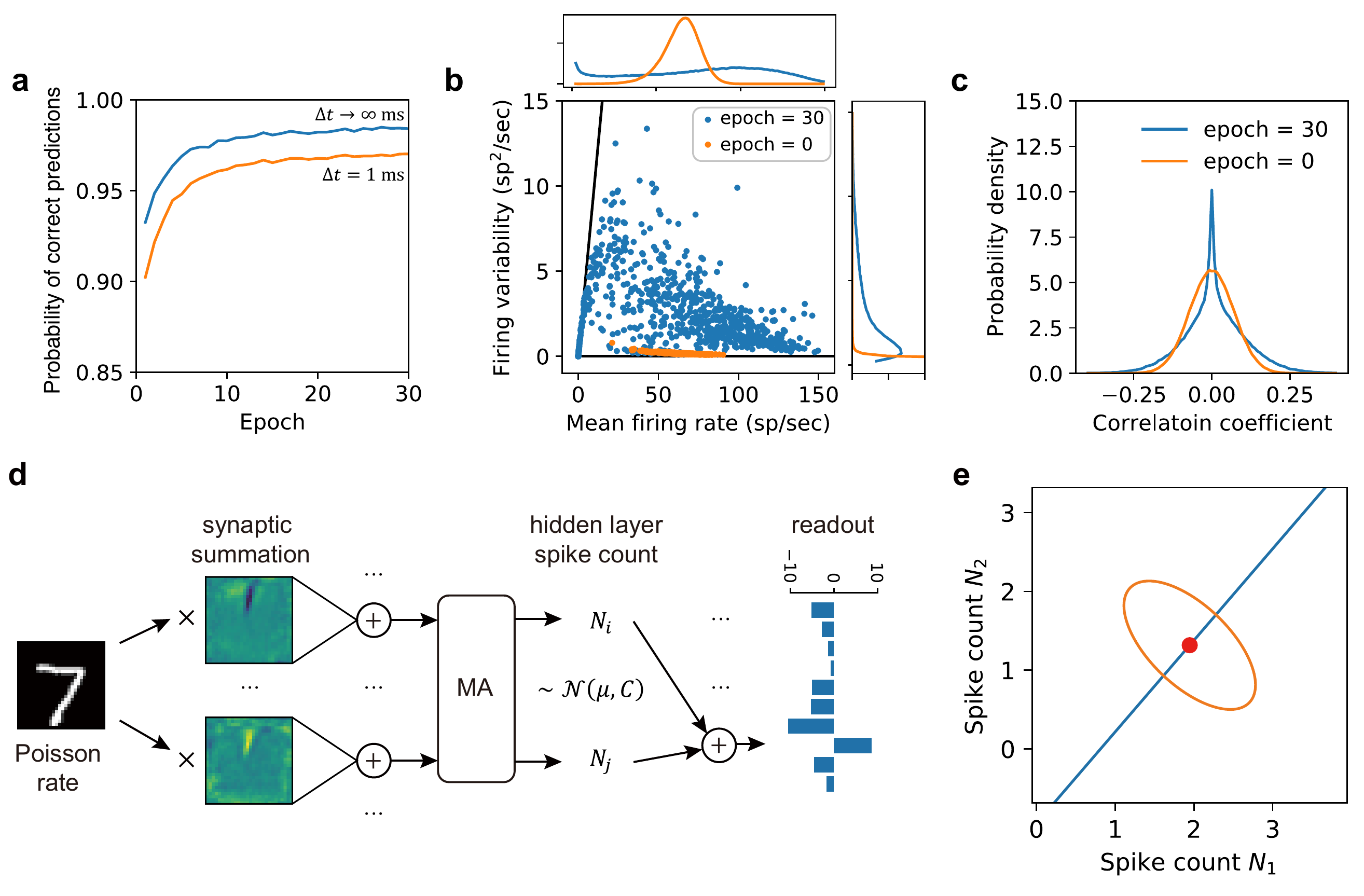}
\caption{\textbf{Moment neural network learning classification task while simultaneously minimizing uncertainty.}  
\textbf{a}, The probability of correct prediction averaged over all samples of the validation set during training for unlimited and limited readout time $\Delta t$; the latter takes into account of trial-to-trial variability. For unlimited readout time ($\Delta t\to\infty$), the accuracy reaches $98.45\%$ at the end of the epochs, comparable to the performance of rate-based artificial neural networks. 
\textbf{b}, Diverse firing variability of hidden layer neurons in response to the input image shown in \textbf{e}, exhibiting both mean-dominant (Fano factor close to zero) and fluctuation-dominant (Fano factor close to one, solid line) activity. Insets: probability densities. 
\textbf{c}, Spike count correlation of the hidden layer neurons exhibit weak correlation whose distribution shows a slower decaying tail after training. 
\textbf{d}, Illustration of non-trivial roles played by correlated variability of a specific pair of hidden layer neurons. An input image represented by independent Poisson spike trains undergoes synaptic summation with anti-correlated weights, leading to anti-correlated neural activity. The final readout is linearly decoded from the hidden layer spike counts. 
\textbf{e}, The mean (dot) and covariance (ellipse) of the spike count of those two neurons over a readout time $\Delta t=100$ ms. In this example, the principal axis of the covariance is orthogonal to the direction of the readout weights (solid line) with respect to the target class, leading to a reduction in the readout variance and simultaneously an increase in the readout mean.
\label{fig:schematic_mnn_vs_snn}}
\end{figure}

Figure~\ref{fig:schematic_mnn_vs_snn}\textbf{a} shows the classification accuracy, measured as the probability of correct prediction [$q_t$ in equation (\ref{eq:general_loss2})] averaged across all images in the validation set, increases with training epochs. When the readout time is infinite [$\Delta t\to\infty$ in equation (\ref{eq:general_loss2})], this simply reflects the fraction of correctly classified samples like in a rate-based artificial neural network. In contrast to rate models, however, the MNN can also express uncertainty (trial-to-trial variability) when the readout time is finite, as reflected by a lower probability of correct prediction. \qi{Note that as $\Delta t$ increases, the probability of correct prediction at finite readout time converges to the theoretical limit. As we will show in a later section, this convergence is exponentially fast with a time scale directly related to the readout uncertainty.}%

In addition, the hidden layer exhibits diverse firing variability consistent with cortical neurons~\citep{softky1993highly,Ponce2013,rosenbaum2017spatial}. Fig.~\ref{fig:schematic_mnn_vs_snn}\textbf{b} shows the neural response to a typical sample image, with each point corresponding to a neuron. The mean firing rate and firing variability of the hidden layer neuron cover a broad range of values, from fluctuation-dominant activity (closer to Fano factor of one, solid line) to mean-dominant activity (closer to Fano factor of zero, x-axis). In contrast, a network with random initialization before training has narrowly distributed firing variability. The pairwise correlations of the hidden layer neurons are also weakly correlated, with both positive and negative values centered around the origin [Fig.~\ref{fig:schematic_mnn_vs_snn}(c)]. This result is consistent with that observed in cortical neurons~\citep{rosenbaum2017spatial}, and also satisfies the assumptions behind the linear response analysis used to derive the correlation mapping in equation (\ref{eq:ma_chi}). Interestingly, we find that the distribution of the correlation coefficients exhibits a longer tail after training. 

To provide an intuitive understanding about the role played by correlated neural variability, we now focus on a specific pair of neurons in the hidden layer and trace the computational steps involved in producing $y_t$, the readout component corresponding to the target class. As shown in Fig.~\ref{fig:schematic_mnn_vs_snn}\textbf{d}, an input image encoded by independent Poisson spikes with $\mu_0=\sigma^2_0$ first undergoes synaptic summation to produce correlated synaptic currents, which in turn elicit neural responses $(\mu_1,C_1)$ in the hidden layer. For the specific pair of neurons shown, the synaptic weights have opposite patterns, resulting in negatively correlated neural responses. Figure~\ref{fig:schematic_mnn_vs_snn}\textbf{e} illustrates the joint distribution of spike count ($\Delta t=100$ ms) for this neuronal pair, with their mean firing rate marked by the dot and their covariance highlighted by the ellipse. Remarkably, the principal axis of the covariance, in this 2D projection, is orthogonal to the line representing the readout weights from these two neurons to the target class [solid line in Fig.~\ref{fig:schematic_mnn_vs_snn}\textbf{e}]. As a result, the readout effectively projects the spike count distribution in the hidden layer along its principal axis, leading to reduced uncertainty in $y_t$.

\subsection{Reconstructing spiking neural network with zero free parameter}%

Because the MNN is analytically derived from the spiking neural network (SNN) model, recovering the SNN from a trained MNN is straightforward. No further post-training optimization or fining tuning is required. First, an input image is encoded into independent Poisson spike trains, which then undergo synaptic summation according to equation (\ref{eq:current}). The synaptic weights and the external currents $I^{\rm ext}_i$ are recovered by absorbing the moment batch normalization into the summation layer of the trained MNN according to equation (\ref{eq:scale_weight}). Finally, the readout $\mathbf{y}(\Delta t)$ is calculated from the spike count $\mathbf{n}(\Delta t)$ over a time window of duration $\Delta t$ according to equation (\ref{eq:snn_readout}). It becomes evident that the readout $\mathbf{y}(\Delta t)$ follows a distribution $p(\mathbf{y}|\mathbf{x};\theta)$ with mean $\mu(\theta)$ and covariance $\tfrac{1}{\Delta t}C(\theta)$ as output by the MNN. The class corresponding to the largest entry in the readout $\mathbf{y}(\Delta t)$ is then taken as the class prediction. %

\begin{figure}[H]
\centering
\includegraphics[width=0.5\textwidth]{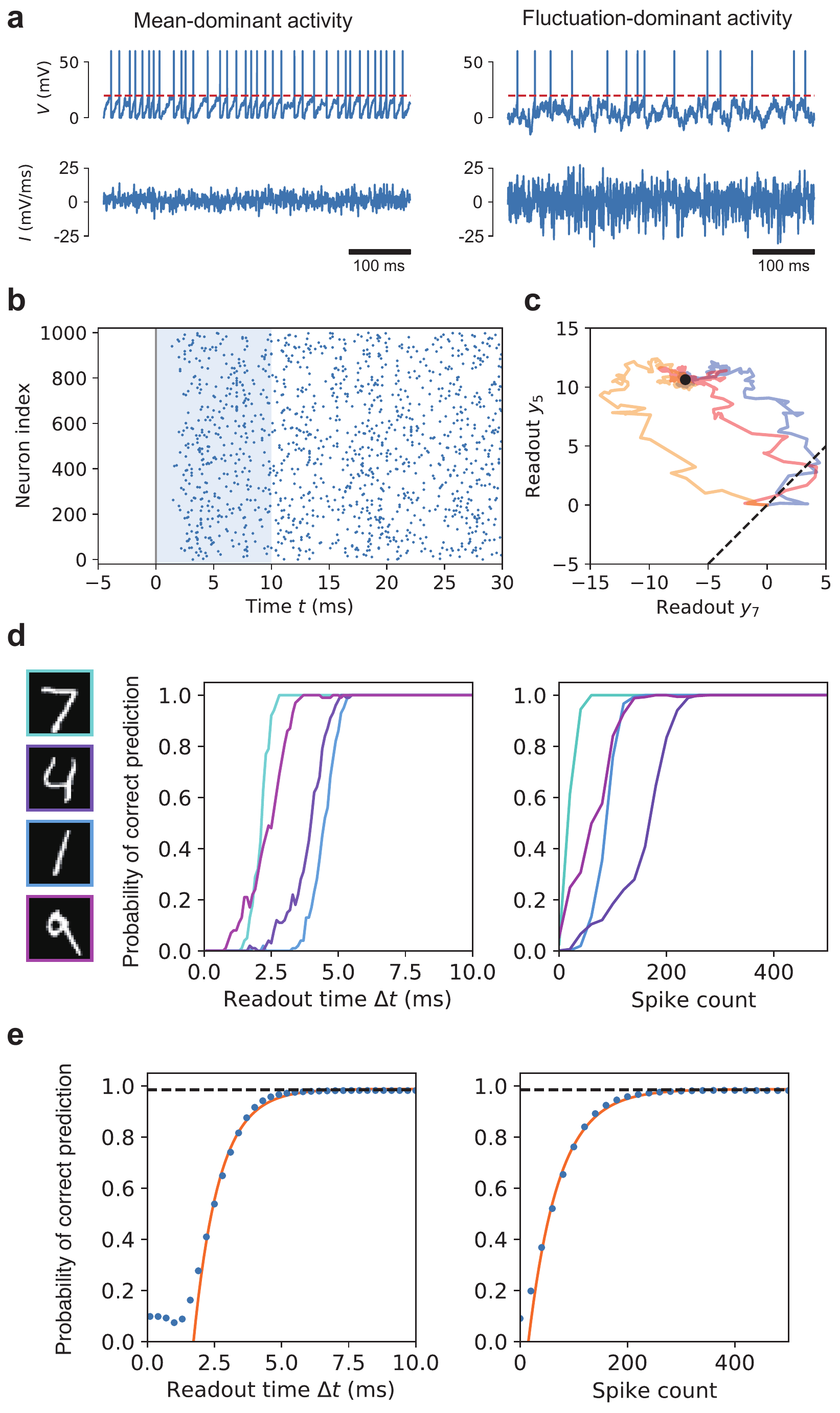}
\caption{
\textbf{Temporal dynamics of stochastic neural computing in spiking neural network.}
\textbf{a}, Membrane potential (upper panel) and synaptic current (lower panel) of two typical hidden layer neurons during one trial of stimulus presentation, each exhibiting mean-dominant (left panel) and fluctuation-dominant (right panel) activity. Dashed line indicates firing threshold. 
\textbf{b}, Raster plot of typical spiking activity of hidden layer neurons during one trial of stimulus presentation. Solid line indicates stimulus onset at $t=0$ ms; shaded region indicates the readout time window $\Delta t$. The membrane potential of all neurons are initialized to zero at $\Delta t=0$ ms. 
\textbf{c}, Two dimensional projection of the readout trajectory $y$ over time for three trials using the same stimulus. The vertical axis is the readout component corresponding to the correct class. Dot indicates theoretical limit of the readout mean as $\Delta t \to\infty$; dashed line indicates the decision boundary. 
\textbf{d}, Probability of correct prediction for a number of input images (left panel) as a function of readout time and population spike count in the hidden layer. Each curve is calculated from $100$ trials with the same stimulus. 
\textbf{e}, Probability of correct prediction averaged over all images of the validation set converges exponentially with the readout time (left panel) as well as the population spike count in the hidden layer (right panel). Dashed lines indicate the theoretical limit of $0.985$ as predicted by the MNN; solid lines represent exponential fits. 
\label{fig:mnist}}
\end{figure}

As consistent with the MNN, the recovered SNN exhibits both mean-dominant and fluctuation-dominant spiking activity as shown in Fig.~\ref{fig:schematic_mnn_vs_snn}\textbf{a}. For a typical neuron with mean-dominant activity, the synaptic current it receives has a positive mean and weak temporal fluctuations. As a result, the sub-threshold membrane potential of the neuron consistently ramps up over time, resulting in spike emission at relatively regular intervals. In contrast, a neuron with fluctuation-dominant activity is largely driven by a synaptic current with large fluctuations even though its mean is close to zero, resulting in spike emission at highly variable intervals. Such diverse firing variability is a key feature of SNC, even if the neuronal model itself is deterministic. The spike raster plot of hidden layer neurons in the SNN in response to an input image [the same as in Fig.~\ref{fig:schematic_mnn_vs_snn}\textbf{d}] is shown in Fig.~\ref{fig:mnist}\textbf{b}. \qi{We find that the MNN accurately captures the distribution of neural activity in the reconstructed SNN (see Fig.~\ref{fig:snn_dist} in Supplementary Information).}

To reveal the temporal dynamics of the readout, we show in Fig.~\ref{fig:mnist}\textbf{c} a 2D projection of the readout trajectories $\mathbf{y}(\Delta t)$ in response to the same image over different trials. When $\Delta t$ is small, the readouts from individual trials are scattered over a wide area, corresponding to a larger trial-to-trial variability. As more spikes are accumulated with increasing $\Delta t$, the trajectory of the readout in a single trial also fluctuates over time, and eventually converges toward the readout mean $\mu$ [marked with the dot in Fig.~\ref{fig:mnist}\textbf{c}] as predicted by the MNN. Since the magnitude of the fluctuations in the readout tends to decrease over time, this may potentially provide a way for the brain to infer confidence during a single trial and potentially an early stopping criterion for decision making. 

To further quantify how task performance depends on readout time, we simulate the SNN over 100 trials for each image in the validation set of MNIST, and calculate the probability of correct prediction [$q_t$ in equation (\ref{eq:general_loss2})] for different input images as the readout time $\Delta t$ increases. As can be seen from the result for four randomly picked images shown in Fig.~\ref{fig:mnist}\textbf{d}, $q_t$ increases with the readout time rapidly and eventually reaches one within around 5 ms, with some images require less time than others. A similar pattern is found for $q_t$ when plotted as a function of spike count (measured by binning individual trials based on the population spike count of hidden layer neurons) which directly reflects the energy cost. 

When averaged over all images, the probability of correct prediction reveals an exponential convergence toward the theoretical limit of $0.9845$ as predicted by the MNN with a short time constant of $\tau=0.95$ ms (left panel in Fig.~\ref{fig:mnist}\textbf{e}). A short burn-in time of around 1 ms is due to the membrane potential being initialized to zero. This rapid convergence results in short decision latency, with an average probability of correct prediction of \qi{0.980 obtained in 6.6 ms}. This is largely due to that the moment cross entropy explicitly takes into account of trial-to-trial variability for finite readout time, so that the neural network learns to improve the rate of convergence without requiring knowledge of precise spike timing. %
A similar exponential convergence for $q_t$ averaged over all images is found with respect to the spike count in the hidden layer, with a decay constant of around 50 spikes. \qi{An average probability of correct prediction of 0.980 can be achieved with 300 spikes in the hidden layer (985 spikes across the entire network if including the input spikes)}. This exceptional energy efficiency is largely due to that a large proportion of the neural population in our model is fluctuation-dominant [Fig.~\ref{fig:schematic_mnn_vs_snn}(a)], with an average firing rate of 50 sp/s per neuron across the hidden layer \qi{and 133 sp/s per neuron in the input layer}. 

\qi{
To gain theoretical insights about this exponential convergence, consider a binary classification problem involving a scalar readout $y=\tfrac{1}{\Delta t}\sum_i w_i n_i(\Delta t)$. Suppose that the correct class corresponds to readout values above a decision boundary $\theta$ and that the readout follows a normal distribution with mean $\mu>\theta$ and variance $\sigma^2/\mathrm{\Delta t}$. Then, the probability of correct prediction is simply 
$q(\Delta t) = \frac{1}{2} \left[1 + \mathrm{erf}\left(\sqrt{\lambda\Delta t}\right)\right]$, where $\lambda=\tfrac{(\mu-\theta)^2}{2\sigma^2}$. Its convergence property for large $\Delta t$ is revealed by its asymptotic expansion
\begin{equation}
q(\Delta t) \sim 1 - \tfrac{1}{\sqrt{4\pi\lambda\Delta t}}
\exp\left (-\lambda\Delta t\right ).
\end{equation}
This analysis shows that as $\Delta t$ increases, the probability of correct prediction $q$ converges exponentially toward one, and as it turns out $\lambda$ plays the role of convergence rate. As $\lambda$ is inversely proportional to the readout variance $\sigma^2$, our analysis provides a quantitative explanation why the inference speed can be improved by minimizing $\sigma^2$. For higher dimensions and decision boundaries with complex shapes, the probability of correct prediction does not have simple expressions but the general principle still applies.
}

To further demonstrate our method, we implement the SNN trained through moment embedding on Intel's Loihi neuromorphic chip and provide benchmarking results in terms of accuracy, energy cost, and latency. See Supplementary Information for details.

\subsection{Analysis of backpropagation in the moment neural network}

\qi{In gradient-based learning, what differentiates one learning algorithm from another is how the gradient $\partial L/\partial \theta$ is estimated. It is therefore instructive to analyze how backpropagation is computed in MNN and how it differs from other methods. Consider a feedforward network with $n$ hidden layers, with each hidden layer consisting the components shown in Fig.~\ref{fig:comp_graph}\textbf{b}. For compactness, denote the output of the $k$-th hidden layer as $\Lambda^{(k)}=(\mu^{(k)},C^{(k)})$ which groups together the mean $\mu^{(k)}$ and covariance $C^{(k)}$ of neural activity into a single vector with $\tfrac{1}{2}N(N+1)$ independent elements. The gradient of the loss function with respect to the model parameters $\theta^{(k)}$, which encompasses the synaptic weights $w^{(k)}$ and the scale and bias parameters $\gamma^{(k)}$, $\beta^{(k)}$, and $C^{{\rm ext},(k)}$, can be expressed as
\begin{equation}
\dfrac{\partial L}{\partial \theta^{(k)}} = \dfrac{\partial L}{\partial \Lambda^{(n)}}
\left [
\dfrac{\partial \Lambda^{(n)}}{\partial \Lambda^{(n-1)}}
\dfrac{\partial \Lambda^{(n-1)}}{\partial \Lambda^{(n-2)}}\dots
\dfrac{\partial \Lambda^{(k+1)}}{\partial \Lambda^{(k)}}
\right ]
\dfrac{\partial \Lambda^{(k)}}{\partial \theta^{(k)}}.
\label{eq:backprop}
\end{equation}
The gradient of each hidden layer can be further written out as
\begin{equation}
\dfrac{\partial \Lambda^{(n)}}{\partial \Lambda^{(n-1)}}
=
\begin{bmatrix}
    \tfrac{\partial \mu^{(n)}}{\partial \bar{\mu}^{(n)}} & \tfrac{\partial \mu^{(n)}}{\partial \bar{C}^{(n)}} \\
    \tfrac{\partial C^{(n)}}{\partial \bar{\mu}^{(n)}} & \tfrac{\partial C^{(n)}}{\partial \bar{C}^{(n)}} \\
\end{bmatrix}
\begin{bmatrix}
    \tfrac{\partial \bar{\mu}^{(n)}}{\partial \hat{\mu}^{(n)}} & \tfrac{\partial \bar{\mu}^{(n)}}{\partial \hat{C}^{(n)}} \\
    \tfrac{\partial \bar{C}^{(n)}}{\partial \hat{\mu}^{(n)}} & \tfrac{\partial \bar{C}^{(n)}}{\partial \hat{C}^{(n)}} \\
\end{bmatrix}
\begin{bmatrix}
    \tfrac{\partial \hat{\mu}^{(n)}}{\partial \mu^{(n-1)}} & 0 \\
    0 & \tfrac{\partial \hat{C}^{(n)}}{\partial C^{(n-1)}} \\
\end{bmatrix},
\label{eq:jacobian}
\end{equation}
where each block, from left to right, represents the Jacobian matrix of the moment activation, the moment batchnormalization, and the synaptic summation, respectively. In practice, the backpropagation is implemented using Pytorch's efficient autograd functionality, with custom gradients for the moment activation~\citep{Qi2022efficient}.} 

\qi{
The general form of backpropagation for the MNN, as revealed by equation (\ref{eq:backprop}) and (\ref{eq:jacobian}), is directly analogous to the backpropagation for rate-based ANNs, as MNN is a natural generalization of ANN to second-order moments. The key difference is that the gradients not only flow through the mean firing rate $\mu$ but also the covariance $C$. To illustrate the idea, consider the Jacobian matrix for the moment activation. There are two ways covariance contributes to the gradient estimation, one of which is through terms directly containing partial derivatives of, or with respect to, covariance (i.e. 
$\tfrac{\partial \mu}{\partial \bar{C}}$, $\tfrac{\partial C}{\partial \bar{\mu}}$, and $\tfrac{\partial C}{\partial \bar{C}}$), and the other through indirect augmentation of the partial derivatives involving only the mean $\tfrac{\partial \mu}{\partial \bar{\mu}}$. In the latter, the role of the input covariance can be interpreted as dynamically controlling the steepness of the mean activation (as can be seen from Fig.~\ref{fig:comp_graph}\textbf{b}), with a larger input variance leading to a smoother gradient. 
}
\qi{
It is also interesting to compare equation (\ref{eq:backprop})-(\ref{eq:jacobian}) with direct gradient estimation based on the temporal dynamics of SNN. 
The most widely adopted approach is backpropagation through time (BPTT) applied directly to the membrane potential dynamics~\citep{Neftci2019IEEE,shrestha2018slayer,10.3389/fnins.2018.00331,10.3389/fnins.2020.00119,Bellec2020}. These methods typically involve surrogate gradients for dealing with discontinuous jumps of membrane potential. However, BPTT and its variants for training SNN are best suited for discrete-time SNN models. For continuous-time SNN models, the neuronal dynamics must first be discretized and unwrapped in time, which may lead to higher cost when finer temporal resolution is required. Another approach to evaluating the gradient is through spike time coding~\citep{BOHTE2002neurocomputing,Wunderlich2021sci_rep}, allowing for exact gradient to be computed. The moment embedding approach presented here provides a way to train continuous-time SNN, without explicitly dealing with temporal dynamics.
}

\subsection{Comparison to existing approaches to training spiking neural networks}
The moment embedding developed in this study leads to a novel approach to training spiking neural networks (SNN) under the principle of stochastic neural computing (SNC). Here, we briefly review existing approaches to training SNNs and compare them to ours. 

Existing approaches to training SNN can largely fall into two categories: ANN-to-SNN conversion and direct SNN training methods~\citep{Roy2019Nature,Pfeiffer2018_deep_learning}. In ANN-to-SNN conversion, a standard artificial neural network (ANN) is pre-trained using traditional backpropagation algorithms, and then converted to an SNN model by mapping the ANN parameters onto SNN. These rate-coded network conversion often results in accuracy loss and relatively high inference latency. Many works have focused on transferring weights from ANN with continuous-valued activation to SNN and setting the firing threshold of neurons layer by layer to achieve near-lossless performance but poor latency \citep{10.3389/fnins.2017.00682,Diehl2015,yan2021near}; other approaches have also been proposed to improve the accuracy while trying to reduce the inference time steps and keep the event-driven sparsity for better energy efficiency \citep{rathi2020enabling}.

Direct approaches to training SNN, on the other hand, use backpropagation-through-time to optimize the parameters of an SNN directly~\citep{Yan2022,Lee2016,Pfeiffer2018_deep_learning}. This requires formulating the SNN as an equivalent recurrent neural network with binary spike inputs. This enables emergent spike-timing codes which convey information with the relative timing between spikes for efficient sparse coding. The main challenge of direct training is due to the discontinuous, non-differentiable nature of spike generation. Diverse methods have been proposed to solve the problem by defining surrogate gradient or spike time coding to enable backpropagation; some examples include spike layer error reassignment (SLAYER)~\citep{shrestha2018slayer}, spatiotemporal backpropagation (STBP)~\citep{10.3389/fnins.2018.00331}, and local online learning via eligibility propagation (e-prop)~\citep{Bellec2020}. In general, direct training approaches tend to achieve fewer spikes, lower latency and energy consumption, but need to overcome difficulties on convergence and performance degradation as network size increases~\citep{Pfeiffer2018_deep_learning}.

\qi{Our approach represents significant advances over previous methods, both conceptual and practical. Conceptually, the moment embedding approach overcomes previous challenges of mapping continuous-valued neural network models to their spiking counterparts by deriving the moment representation of spiking neural activity from first principles. As a consequence, the MNN can serve as an effective proxy for training SNN, not by \textit{conversion} but by \textit{reconstruction}, as the MNN represents a \textit{tight} approximation to SNN under the view of probabilistic coding. The practical benefit this brings about is that no extra parameters are introduced both before and after training. As illustrated in Fig.~\ref{fig:comp_graph}\textbf{b}, the only trainable parameters in the MNN are the synaptic weights, and the scaling factors and biases in the moment batch normalization. Notably, the dimensionality of these trainable parameters are identical to its ANN-counterpart. We find that the performance of MNN is comparable to its ANN counterpart under the same network structures (see table~\ref{tab:extra_experiments} in Supplementary Information). In contrast, ANN-to-SNN conversion requires extra parameters or various post-training optimization techniques to eliminate conversion error, such as manually adjusting the firing threshold and the scaling of firing rate at the input layer~\citep{Diehl2015}, modifying or adding trainable parameters to the activation function~\citep{ding2021optimal,yan2021near,Hunsberger2015arxiv}, weight scaling~\citep{Kim22811} and weight normalization~\citep{Diehl2015,10.3389/fnins.2017.00682}. We point out that our method does not prohibit post-training optimizations and both techniques can be applied in tandem in the future to further boost the performance of the SNN. Additionally, ANN-to-SNN conversion is often criticized for its reliance on rate coding that leads to slow convergence in the converted SNN. It has been suggested that post-training weight normalization could improve the convergence~\citep{Diehl2015}. Here, by generalizing ANN to second-order moments, the MNN enables the representation of uncertainty and subsequently reduction of the inference time through manipulating correlated neural variability. Our approach thus provides not only an alternative way to train SNN but also insights to how it operates through the lens of stochastic neural computing. See table~\ref{tab:model_comparison} in Supplementary Information for a quantitative comparison of performance between different methods.}

\qi{A number of studies have explored how noise can be exploited for various benefits during learning by using simplified probabilistic spiking neural network models~\cite{Ma2023patterns,Jang2019,Dutta2022nat_comm}. By modeling spike generation as a Bernoulli process with firing probability depending on the neuron's input, it has been shown that a three-factor learning rule can be recovered from a probability learning objective~\cite{Jang2019}. Using simplified binary neural network with stochastic synapses, it has been shown that multiplicative synaptic noise can lead to intrinsic weight normalization during learning~\citep{Detorakis2019nips}. However, these simplified models are limited to uncorrelated neural spiking activity. By exploiting the spiking neuron's innate ability to nonlinearly couple signal and noise, our work demonstrates for the first time the learning and implementation of stochastic neural computing in spiking neural networks with correlated neural variability, and highlights how noise correlation can be exploited to reduce prediction uncertainty and in turn enhance inference speed.}

 \section{Discussion}

The general framework of stochastic neural computing (SNC) based on moment embedding as proposed in this work has a number of advantages. First, the moment embedding approach provides an effective finite-dimensional parameterization of joint probability distributions of neural population activity up to second-order statistical moments, through which probabilistic neural operations can be efficiently computed. Second, derived from spiking neural models on a mathematically rigorous ground, the moment embedding faithfully captures the nonlinear coupling of correlated neural variability and the propagation of neural correlation across large populations of spiking neurons. Lastly, the differentiability of the moment mapping enables backpropagation for gradient-based learning, leading to a new class of deep learning model referred to as the moment neural network (MNN).

The MNN naturally generalizes standard artificial neural network (ANN) in deep learning to second-order statistical moments and provides a conceptual link between biological SNN and ANN, and between spike-time coding and rate coding. Although the example presented here only considers a feedforward architecture, the moment embedding approach can be used to systematically generalize many of the known deep learning architectures, such as convolutional and recurrent neural networks, to second-order statistical moments. %
The moment embedding approach also provides an alternative algorithm for training SNNs based on probabilistic codes. Unlike ANN-to-SNN conversion methods in the deep learning literature, which requires extensive post-training optimization such as threshold balancing to mitigate performance loss caused by conversion~\citep{Diehl2015,Hunsberger2015arxiv,ding2021optimal,10.3389/fnins.2017.00682,Kim22811,yan2021near}, the parameters trained through the moment embedding can be used directly to recover the SNN without further fining tuning. %

The proposed framework of SNC incorporates uncertainty into the learning objective and further enables direct manipulations of correlated neural variability in a task-driven way. Specifically, the moment embedding approach enables end-to-end learning of arbitrary probabilistic computation tasks, and provides significant advantage over conventional handcrafted approach to constructing neural circuit model for probabilistic neural computation which often rely on prior assumptions regarding the specific form of neural code or simplifications for facilitating theoretical analysis~\citep{Pouget2013a,Ma2014a}. Our theory of SNC also emphasizes uncertainty representation through stochastic processes of neural spike trains, through which signal and noise are processed concurrently rather than through different channels such as in a variational auto-encoder.

\qi{While second-order moments do not provide the complete information about a joint probability distribution, our approach represents a major leap compared to uncorrelated Poisson or binary neurons commonly assumed in the literature. It is arguably the only viable way to calculate the propagation of correlated neural activity without resorting to full simulation. 
As we have shown, the spike count distributions obtained from simulations of the SNN well match to that predicted by the MNN (Fig.~\ref{fig:snn_dist}), indicating the validity of using second-order moments for modeling stochastic neuronal dynamics. 
The validity of this approach is also supported by experimental and theoretical evidence. It has been found that the collective behavior exhibited by the neural population in the vertebrate retina is well accounted for by a model that captures the observed pairwise correlations but assumes no higher-order interactions~\citep{schneidman2006weak}. A theoretical study investigates the effect of higher-order moments in a stochastic binary neural network model~\citep{Dahmen2019pnas} and has found that in a weakly correlated state, the contributions from higher-order moments to the network dynamics are significantly smaller than pairwise interactions. This suggests that the first two moments are quite sufficient for modeling the stochastic dynamics of neural activity.}

The approach developed in this paper also has broader implications to stochastic computing, which has been proposed as an alternative computing architecture for approximate computation with better error tolerance and energy efficiency~\citep{gaines1969stochastic,Alaghi2018}. However, designing stochastic computing circuits for arbitrary functions remains a major challenge. Our method indicates a solution to this problem by training SNNs for implementing spike-based SNC. Combined with advances in neuromorphic hardware~\citep{Davies2021,8259423,Dutta2022nat_comm}, the principle of SNC could lead to a future generation of brain-inspired computing architecture.

\section{Methods}\label{sec11}
\subsection{Leaky integrate-and-fire neuron model}
The membrane potential dynamics of a leaky integrate-and-fire (LIF) neuron is described by%
\begin{equation}
\dfrac{dV_i}{dt}= -LV_i(t) + I_i(t),
\label{eq:LIF}
\end{equation} 
where the sub-threshold membrane potential $V_i(t)$ of a neuron $i$ is driven by the total synaptic current $I_i(t)$ and $L=0.05$ \si{\per\milli\second} is the leak conductance. When the membrane potential $V_i(t)$ exceeds a threshold $V_{\rm th}=20$ \si{\milli\volt} a spike is emitted, as represented with a Dirac delta function. Afterward, the membrane potential $V_i(t)$ is reset to the resting potential $V_{\rm res}=0$ mV, followed by a refractory period $T_{\rm ref}=5$ ms. The synaptic current takes the form
\begin{equation}
I_i(t)= \sum_{j}w_{ij}S_j(t)+I_i^{\rm ext}(t),
\label{eq:current}
\end{equation}
where $S_j(t)=\sum_k \delta(t-t^k_j)$ represents the spike train generated by pre-synaptic neurons.

A final output $\mathbf{y}$ is readout from the spike count $\mathbf{n}(\Delta t)$ of a population of spiking neurons over a time window of duration $\Delta t$ as follows
\begin{equation}
y_i(\Delta t) = 
\frac{1}{\Delta t}\sum_{j} w_{ij}n_{j}(\Delta t) + \beta_i, %
\label{eq:snn_readout}
\end{equation}
where $w_{ij}$ and $\beta_i$ are the weights and biases of the readout, respectively. One property of the readout is that its variance should decrease as the readout time window $\Delta t$ increases. %

\subsection{Moment embedding for the leaky integrate-and-fire neuron model}

\qi{The first- and second-order moments of spiking neural activity, which we refer to as the mean firing rate and the firing co-variability, are defined as}
\begin{equation}
\mu_i = \lim_{\Delta t\to\infty} \dfrac{\mathbb{E}[n_i(\Delta t)]}{\Delta t}, 
\label{eq:def_mean}
\end{equation}
and
\begin{equation}
C_{ij} = \lim_{\Delta t\to\infty} \dfrac{{\rm Cov}[n_i(\Delta t),n_j(\Delta t)]}{\Delta t}, 
\label{eq:def_cov}
\end{equation}
respectively, where $n_i(\Delta t)$ is the spike count of neuron $i$ over a time window $\Delta t$. \qi{The limits signify that we consider the statistics of the system after it has reached stationarity. For stationary processes, the trial-wise expectation of spike count over finite time windows does not depend on $\Delta t$, that is, $\mathbb{E}[n_i(\Delta t)] = \mu_i\Delta t$ holds for all $\Delta t$. The covariance of spike count, however, always depend on the time window (as does Fano factor~\citep{Rajdl2020}), even if the system is stationary. In practice, we use ${\rm Cov}[n_i,n_j]\approx C_{ij}\Delta t$ as an asymptotic approximation, which becomes increasingly accurate with larger $\Delta t$. %
}

For the LIF neuron model [equation (\ref{eq:LIF})], the statistical moments of the synaptic current is equal to~\citep{Feng2006,LU2010913}%
\begin{numcases}{}%
\hat{\mu}_i = \textstyle\sum_kw_{ik}\mu_k + \hat{\mu}_i^{\rm ext},\label{eq:sum_mean}
\\
\hat{C}_{ij}=\textstyle\sum_{kl} w_{ik}C_{kl}w_{jl} + \hat{C}_{ij}^{\rm ext},\label{eq:sum_cov}
\end{numcases}
where $w_{ik}$ is the synaptic weight and $\hat{\mu}_i^{\rm ext}$ and $\hat{C}_{ij}^{\rm ext}$ are the mean and covariance of an external current, respectively. 
Note that from equation (\ref{eq:sum_cov}), it becomes evident that the synaptic current are correlated even if the pre-synaptic spike trains are not. 
Next, the first- and second-order moments of the synaptic current is mapped to that of the spiking activity of the post-synaptic neurons. For the LIF neuron model, this mapping can be obtained in closed form through a mathematical technique known as the diffusion approximation~\citep{Feng2006,LU2010913} as
\begin{numcases}{}
\mu_i = \phi_\mu(\bar{\mu}_i,\bar{\sigma}_i),\label{eq:ma_mu}\\
\sigma_i = \phi_\sigma(\bar{\mu}_i,\bar{\sigma}_i),\label{eq:ma_sig}\\
\rho_{ij} = \chi(\bar{\mu}_i,\bar{\sigma}_i)\chi(\bar{\mu}_j,\bar{\sigma}_j)\bar{\rho}_{ij},\label{eq:ma_chi}
\end{numcases}
where the correlation coefficient $\rho_{ij}$ is related to the covariance as $C_{ij}=\sigma_i\sigma_j\rho_{ij}$. In this paper, we refer this mapping given by equation (\ref{eq:ma_mu})-(\ref{eq:ma_chi}) as the moment activation.

The functions $\phi_\mu$ and $\phi_\sigma$ together map the mean and variance of the input current to that of the output spikes according to~\citep{Feng2006,LU2010913}
\begin{numcases}{\phi: (\bar{\mu},\bar{\sigma}^2) \mapsto(\mu,\sigma^2),~~~~}
\mu =  \dfrac{1}{T_{\rm ref} + \tfrac{2}{L}\int_{I_{\rm lb}}^{I_{\rm ub}} g(x) dx},\label{eq:mu}
\\
\sigma^2 = \tfrac{8}{L^2}\mu^3\textstyle\int_{I_{\rm lb}}^{I_{\rm ub}} h(x) dx,\label{eq:sigma}
\end{numcases}
where $T_{\rm ref}$ is the refractory period with integration bounds $I_{\rm ub}(\bar{\mu},\bar{\sigma}) = \tfrac{V_{\rm th}L-\bar{\mu}}{\sqrt{L}\bar{\sigma}}$ and
$I_{\rm lb}(\bar{\mu},\bar{\sigma}) = \tfrac{V_{\rm res}L-\bar{\mu}}{\sqrt{L}\bar{\sigma}}$. The constant parameters $L$, $V_{\rm res}$, and $V_{\rm th}$ are identical to those in the LIF neuron model in equation (\ref{eq:LIF}). The pair of Dawson-like functions $g(x)$ and $h(x)$ appearing in equation (\ref{eq:mu}) and equation (\ref{eq:sigma}) are $g(x)=e^{x^2}\int_{-\infty}^x e^{-u^2}du$ and $h(x)=e^{x^2}\int_{-\infty}^x e^{-u^2}[g(u)]^2du$. The function $\chi$, which we refer to as the linear perturbation coefficient, is equal to $\chi(\bar{\mu},\bar{\sigma})=\tfrac{\bar{\sigma}}{\sigma}\tfrac{\partial\mu}{\partial\bar{\mu}}$ and it is derived using a linear perturbation analysis around $\bar{\rho}_{ij}=0$~\citep{LU2010913}. This approximation is justified as pairwise correlations between neurons in the brain are typically weak. An efficient numerical algorithm is used for evaluating the moment activation and its gradients~\citep{Qi2022efficient}.

\qi{
Finally, the moments of the readout $y_i$ [in equation (\ref{eq:snn_readout})] can be calculated as
\begin{numcases}{}
\mathbb{E}[y_i] = \sum_{j} w_{ij}\mu_j + \beta_i,\label{eq:mnn_readout_mean}\\
\mathrm{Cov}[y_i,y_j] = \dfrac{1}{\Delta t}\sum_{k,l} w_{ik}C_{kl}w_{jl} ,\label{eq:mnn_readout_cov}
\end{numcases}
where $w_{ij}$ and $\beta_i$ are the weights and biases of the readout, respectively. Here, $\mu_j$ and $C_{ij}$ denote the mean firing rate and firing covariability as calculated by the moment activation [equation (\ref{eq:ma_chi})]. Note that in deriving equation (\ref{eq:mnn_readout_cov}), we have used the approximation $\mathrm{Cov}[n_i,n_j]\approx C_{ij}\Delta t$. Unlike the readout mean, the readout covariance depends on the readout duration $\Delta t$, showing one of the striking differences between deterministic and stochastic neural computing.
} %
\subsection{Moment batch normalization} 

The moment batch normalization for the input mean $\hat{\mu}_i$ is
\begin{equation}
\bar{\mu_i} = \dfrac{\hat{\mu}_i-\mathbb{E}[\hat{\mu}_i]}{\sqrt{\nu_i+\epsilon}}\gamma_i+\beta_i,
\label{eq:bn_mean}
\end{equation}
where $\mathbb{E}[\hat{\mu}_i]$ is the mean computed over samples within a mini-batch and $\nu_i$ is a normalization factor. The bias $\beta_i$ and scaling factor $\gamma_i$ are trainable parameters, similar to that in the standard batch normalization. The key difference from the standard batch normalization for firing rate model is the normalization factor which must accommodate the effect of input fluctuations. In this study, we propose the following form of normalization factor, $\nu_i = {\rm Var}[\hat{\mu}_i]+\mathbb{E}[\hat{\sigma}^2_i]$,
which involves the expectation of the input variance in addition to the variance of the input mean. In fact, by invoking the law of total variance, it can be shown that this particular choice of normalization factor can be interpreted as the variance of the total synaptic current $I_i(t)$ [equation (\ref{eq:current})] in the corresponding SNN, that is, ${\rm Var}[\hat{\mu}_i]+\mathbb{E}[\hat{\sigma}^2_i]={\rm Var}[I_i(t)]$, where the variance on the right-hand side is evaluated across the mini-batch as well as time. Note that the standard batch normalization used in rate-based ANN corresponds to the special case of equation (\ref{eq:bn_mean}) when the input current $I_i(t)$ is constant, that is, when $\hat{\sigma}^2_i=0$.

The moment batch normalization for the input covariance $\hat{C}_{ij}$ enforces the same normalization factor $\nu_i$ and trainable $\gamma_i$ as used in equation (\ref{eq:bn_mean}) but without centering. The shared normalization factor and trainable factor allow the moment batch normalization to be absorbed into the synaptic weights after training is complete, thereby preserving the link to the underlying spiking neural network. This leads to
\begin{equation}
\bar{C}_{ij} = 
\dfrac{\hat{C}_{ij}}{
\sqrt{(\nu_i+\epsilon)(\nu_j+\epsilon)}
}\gamma_i\gamma_j
+C_{ij}^{\rm ext},
\label{eq:bn_cov}
\end{equation}
where $C_{ij}^{\rm ext}$ represents the covariance of an external input current. To ensure symmetry and positive semi-definiteness of the covariance matrix, we set $C^{\rm ext} = A^TA$ with the matrix $A$ being a trainable parameter with the same size as $C^{\rm ext}$. Alternatively, for independent external input current, we set $C^{\rm ext}_{ij} = \delta_{ij}(\sigma^{\rm ext}_i)^2$, with $\sigma_i^{\rm ext}$ being trainable parameters. In practice, the computation of equation (\ref{eq:bn_cov}) can be quite cumbersome and one way to significantly simplify this step, with some reduced flexibility, is to consider the special case where the external input covariance $C_{ij}^{\rm ext}$ is zero. Under this scenario, we only need to apply batch normalization to the variance and pass directly the off-diagonal entries via the correlation coefficient, $\bar{\rho}_{ij}=\hat{\rho}_{ij}$.  Similar to the standard batch normalization, the input mean and input variance over minibatch are replaced by the running mean and running variance during the validation phase. A schematic diagram showing the moment batch normalization is shown in Fig.~\ref{fig:comp_graph}.

A practical benefit of the moment batch normalization is that it simplifies parameter initialization before training as we can initialize of the parameters $(\beta_i,\gamma_i,C_{ij}^{\rm ext})$ to appropriate values so that the total post-synaptic current $(\bar{\mu}_i,\bar{\sigma}_i)$ is always within a desired working regime, regardless of the task or the input sample.
\subsection{Moment loss functions}
Assuming a Gaussian-distributed readout and substituting its probability density
\[
p(\mathbf{y}|\mathbf{x};\theta)= 
\tfrac{1}{\sqrt{ \det(2\pi C/\Delta t) }}\exp\left[
-\tfrac{1}{2}(\mathbf{y}-\mu)^TC^{-1}(\mathbf{y}-\mu)\Delta t
\right]
\]
into each of equation (\ref{eq:general_loss1})-(\ref{eq:general_loss2}) lead to the following objective functions expressed in terms of the second-order statistical moments of the readout. \qi{Here, we use $\mu$ and $C/\Delta t$ to denote the readout mean $\mathbb{E}[\mathbf{y}]$ and readout covariance $\mathrm{Cov}[\mathbf{y},\mathbf{y}]$ in equation (\ref{eq:mnn_readout_mean})-(\ref{eq:mnn_readout_cov}).}

For regression problems, the principle of maximum likelihood leads to
\begin{equation}
L(\theta)%
=(\mu-\mathbf{y}^*)^T C^{-1}(\mu-
\mathbf{y}^*)\Delta t + \log \det (\tfrac{2\pi}{\Delta t} C),
\label{eq:mmse}
\end{equation}
where $\mathbf{y}^*$ represents the readout target and $t$ represents matrix transpose. We refer this loss function as the moment mean-squared error (MMSE) loss. 
In practice, a small positive value (representing a constant external background noise) is added to the diagonal entries of $C$ to avoid numerical instability during matrix inversion. 

For classification problems, class prediction is obtained by taking the class label $i$ corresponding to the largest entry of $\mathbf{y}$. Since there is no simple analytical expression for the probability of correct predictions $q_i(\theta) = \int p(\mathbf{y}\vert\theta)\mathbf{1}_{D_i}(\mathbf{y})d\mathbf{y}$ in high dimensions, we use a finite-sample approximation such that 
$q_i\approx \tfrac{1}{N}\sum_{n=1}^N \mathbf{1}_{D_i}(\mathbf{y}^n)$, with $\mathbf{y}^n$ being a multivariate normal random variable with mean $\mu$ and covariance $\tfrac{1}{\Delta t}C$. To generate the random samples, we perform Cholesky decomposition $C=LL^T$ to express $\mathbf{y}^n$ as $\mathbf{y}^n=L\mathbf{z}^n+\mu$, where $\mathbf{z}^n$ is an uncorrelated unit normal random variable. Importantly, the Cholesky decomposition $L$ is differentiable with respect to $C$, allowing for backpropagation to be implemented. Next, to solve the non-differentiability of the indicator function, we approximate it with the soft-max function 
$\mathbf{1}_{D_i}(\mathbf{y})\approx \sigma_i(\mathbf{y};\beta)= \exp(\beta y_i)/\sum_i\exp(\beta y_i)$, where $\beta$ is a steepness parameter such that $\sigma_i(\mathbf{y};\beta)\to\mathbf{1}_{D_i}(\mathbf{y})$ as $\beta\to\infty$. Combining all these steps we obtain the following generalized cross-entropy loss
\begin{equation}
H(\theta)\approx -\log\sum_{n=1}^N \sigma_t(\tfrac{1}{\sqrt{\Delta t}}L\mathbf{z}^n+\mu;\beta),
\label{eq:loss_CE}
\end{equation}
which we refer to as the moment cross-entropy (MCE) loss (here $t$ denotes target class). Note that the standard cross-entropy loss commonly used in deep learning corresponds to a special case of equation (\ref{eq:loss_CE}) when the readout time is unlimited, that is, when $\Delta t\to\infty$. %
\subsection{Recovering synaptic weights in spiking neural network}

The synaptic weights $\tilde{w}_{ij}$ and the moments of the external currents $I^{\rm ext}_i(t)$ are recovered by absorbing the moment batch normalization into the summation layer of the trained MNN according to the formulae
\begin{equation}
\mu_i^{\rm ext}=\beta_i-\tfrac{\gamma_i}{\sqrt{\nu_i+\epsilon}}\mathbb{E}[\hat{\mu}_i],
\label{eq:shift_current}
\end{equation}
\begin{equation}
\tilde{w}_{ij} = \tfrac{\gamma_i}{\sqrt{\nu_i+\epsilon}} w_{ij},
\label{eq:scale_weight}
\end{equation}
where $w_{ij}$ is the synaptic weight of the summation layer in the trained MNN; the quantities $\mathbb{E}[\hat{\mu}_i]$, $\nu_i= {\rm Var}[\hat{\mu}_i]+\mathbb{E}[\hat{\sigma}^2_i]$, $\beta_i$ and $\gamma_i$ are the running mean, running variance, bias and scaling factor in the moment batch normalization [equation (\ref{eq:bn_mean})-(\ref{eq:bn_cov})]. The covariance of the external current $C_{ij}^{\rm ext}$ is the same as that in equation (\ref{eq:bn_cov}). The external current $I^{\rm ext}_i(t)$ to the spiking neural network can therefore be reconstructed as a Gaussian white noise with mean $\mu_i^{\rm ext}$ and covariance $C_{ij}^{\rm ext}$, and in turn be fed into the LIF neuron model in equation (\ref{eq:current}). No further post-training optimization or fining tuning is required during this reconstruction procedure.

\subsection{Model setup for training}

We train the moment neural network on the MNIST dataset which contains 60000 images for training and 10000 images for validation. The model consists of an input layer, a hidden layer and a readout layer. For this task, the number of neurons is 784 for the input layer, 1000 for the hidden layer, and 10 for the readout. 
For the input layer, a Poisson-rate encoding scheme is used such that neurons in the input layer emits independent Poisson spikes with rates proportional to the pixel intensity $\mathbf{x}$, that is, $\mu_0 = \sigma^2_0 = \alpha \mathbf{x}$, where $\alpha$ is the stimulus transduction factor set to be $\alpha=1$ spikes per ms, and the correlation coefficient $\rho_{ij}=0$ for $i\neq j$. The hidden layer involves synaptic summation [equation (\ref{eq:sum_mean})-(\ref{eq:sum_cov})], followed by moment batch normalization [equation (\ref{eq:bn_mean})-(\ref{eq:bn_cov})] and then by moment activation [equation (\ref{eq:ma_mu})-(\ref{eq:ma_chi})]. The readout mean $\mu$ and covariance $\tfrac{1}{\Delta t}C$ are calculated using equation (\ref{eq:sum_mean})-(\ref{eq:sum_cov}) where the readout time is set to be $\Delta t=1$ ms. %
The moment cross entropy loss [equation (\ref{eq:loss_CE})] is used to train the network. with the number of random samples set to be $n=1000$ and the steepness parameter to $\beta=1$ during training. %
The model is implemented in Pytorch and trained with stochastic gradient descent (AdamW). Gradients are evaluated using Pytorch's autograd functionality, except for the moment activation in which custom gradients for equation (\ref{eq:ma_mu})-(\ref{eq:ma_chi}) are used~\citep{Qi2022efficient}. %

\section*{Code Availability}
The code for simulating and training the moment neural network (MNN) model is available without restrictions on Github (https://github.com/BrainsoupFactory/moment-neural-network).

\section*{Acknowledgments}
Supported by STI2030-Major Projects (No. 2021ZD0200204); supported by Shanghai Municipal Science and Technology Major Project (No. 2018SHZDZX01), ZJ Lab, and Shanghai Center for Brain Science and Brain-Inspired Technology; supported by the 111 Project (No. B18015).

\section*{Author Contributions}

Conceptualization: Y.Q. and J.F.; Methodology: Y.Q., W.L. and J.F.; Investigation: Y.Q., Z.Z. and Y.W.; Software: Y.Q., Z.Z. and Y.W.; Resources and deployment supporting: Z.W. and L.C.; Visualization: Y.Q. and Z.Z.; Writing — original draft: Y.Q. and Z.Z.; Writing — review \& editing: Y.Q., Z.Z., Z.W., L.C., J.Z., W.L., and J.F.; Supervision: J.F.

\section*{Competing interests}
The authors declare no competing interests.

\section*{Materials \& Correspondence}
Correspondence and requests for materials should be addressed to J.F.

\newpage

\begin{flushleft}
{\Large
\textbf\newline{{\bf Supplementary Information}}%
\newline{}\newline{Toward stochastic neural computing}
}
\newline
\\
Yang Qi\textsuperscript{1,2,3}, 
Zhichao Zhu\textsuperscript{1}, 
Yiming Wei\textsuperscript{1}, 
Lu Cao\textsuperscript{4},
Zhigang Wang\textsuperscript{4},
Jie Zhang\textsuperscript{1,2},
Wenlian Lu\textsuperscript{1,2},
Jianfeng Feng\textsuperscript{1,2,*}
\\
\bigskip
\it{1} Institute of Science and Technology for Brain-Inspired 
Intelligence, Fudan University, Shanghai 200433, China
\\
\it{2} Key Laboratory of Computational Neuroscience and Brain-Inspired 
Intelligence (Fudan University), Ministry of Education, China
\\
\it{3} MOE Frontiers Center for Brain Science, Fudan University, Shanghai 200433, China
\\
\it{4} Intel Labs China, Beijing, 100190, China

\bigskip 
\end{flushleft}

\setcounter{section}{0}
\renewcommand{\thesection}{S\arabic{section}}
\setcounter{equation}{0}
\renewcommand{\theequation}{S\arabic{equation}}
\setcounter{figure}{0}
\renewcommand{\thefigure}{S\arabic{figure}}
\setcounter{table}{0}
\renewcommand{\thetable}{S\arabic{table}}

\setcounter{page}{1} %

\section{Spiking neural network implementation on Loihi}

As a research chip, Loihi is designed to support a broad range of SNNs with sufficient scale, performance, and features to deliver competitive results compared to state-of-the-art contemporary computing architectures~\citep{Davies2021,8259423,renner2021backpropagation}. Loihi harnesses insights from biological neural systems to build electronic devices and to realize biological computing principles efficiently. Specifically, Loihi is a digital, event-driven asynchronous circuit chip using discrete-time computational model. Each Loihi chip contains 128 neuromorphic cores and 3 low-power embedded x86 CPU cores. Every neuromorphic core has up to 1K leaky integrate-and-fire (LIF) neurons, implementing a total of 128K neurons and 128M synaptic connections per chip for spike-based neural computing. Embedded CPU cores are mainly used for management as well as bridging between conventional and neuromorphic hardware. Single Loihi chip could show orders of magnitude improvements in terms of low power consumption and fast processing speed on certain tasks compared with conventional hardware.

Loihi implements a discretized version of the LIF neuron model as follows 
\begin{align}
     V_i(t) & = (1 - \tfrac{1}{\tau_{v}} ) V_i(t-1)  + I_i(t) + \beta, \\
     I_i(t) & = (1 - \tfrac{1}{\tau_{u}} ) I_i(t-1)  + \sum_j w_{ij} s_j(t),
     \label{eq:loihi_model}
\end{align}
where $V_i(t)$ and $I_i(t)$ represent the membrane potential and synaptic current received by the $i$th neuron at a discrete time step $t$, respectively, and $s_j(t)$ represents binary spike trains of pre-synaptic neurons. The quantities $\tau_v$, $\tau_u$, and $\beta$ represent the membrane time constant, synaptic time constant, and bias current, respectively. 

The following parameter setting is used to match the discrete LIF model [equation (\ref{eq:loihi_model})] implemented by Loihi to the continuous LIF model [equation (\ref{eq:LIF})] in the main text. First, the synaptic current time constant $\tau_u$ is set to be infinite so that the synaptic current becomes a weighted sum of input spikes. Second, we use exact solution of the subthreshold membrane potential rather than the Euler scheme for numerical integration to accommodate larger simulation time step $\delta t$, which is often desirable for saving energy on hardware. As a result, the membrane time constant is set to be $\frac{1}{\tau_v} = 1 - e^{-L\delta t}$ and the bias current is set to be $\beta_i = \tfrac{1}{L}(1-e^{-L\delta t})I_i^{\rm ext}$, where $I_i^{\rm ext}$ is the external input current in the continuous model. By default, we set $L = 0.05$ \si{\per\milli\second} and $\delta t = 1$ ms.

Since Loihi only supports 9-bit signed integer representation of synaptic weights taking values in the range $[-256, 254]$, appropriate rescaling is needed to map the synaptic weights from physical units to this 9-bit representation. Specifically, a factor of $k = \frac{254}{\max(|\hat{w}^{\rm MNN}|)}$ is used to scale the synaptic weight, the membrane potential and the bias currents, which are rounded down to integer values afterward. %

One nuance regarding the hardware implementation is how to emulate the linear readout [equation (\ref{eq:snn_readout})] used in the continuous model. This is achieved with an output layer consisting of perfect integrator neurons ($L=0$) with infinite firing threshold. Under this setting, the linear readout $y$ is equivalent to the membrane potential in the output layer divided by the readout time $T$, which can be recorded by the hardware. The model prediction is based on the output neuron with the highest membrane potential. 

\section{Benchmarking results on Loihi}

To demonstrate, we implement the spiking neural network (SNN) trained using our method to perform MNIST image classification task on Intel's neuromorphic chip Loihi. Unless otherwise specified, the network structure is the same as in the main text. We first perform off-chip training of the moment neural network and then transfer the trained synaptic weights to SNN using equation (\ref{eq:shift_current})-(\ref{eq:scale_weight}) in Methods. For this purpose, we use the moment loss function [equation (\ref{eq:loss_CE})] with $\Delta t$ set to be infinite as this setting gives the best result on Loihi. Then, the model parameters are rescaled and quantized to satisfy Loihi's hardware constraints as discussed above. For hardware implementation, the stimulus transduction factor is set to be $\alpha=0.1$ sp/ms for encoding training sample (see Method in main text). This further limits the overall number of spikes and thus energy consumption.

Performance of hardware implementation of an SNN is typically characterized by a number of key metrics including accuracy, energy consumption, and latency (also known as delay). Accuracy measures the fraction of correctly predicted samples in the dataset averaged across trials. Energy consumption measures in joule the dynamic energy consumed by the neuromorphic cores per inference, whereas delay measures the elapsed wall time per inference. Both energy consumption and delay are measured by the on-board probe of Loihi Nahuku-32 board using NxSDK version 2.0. 

\begin{figure}
    \centering
    \includegraphics[width=\textwidth]{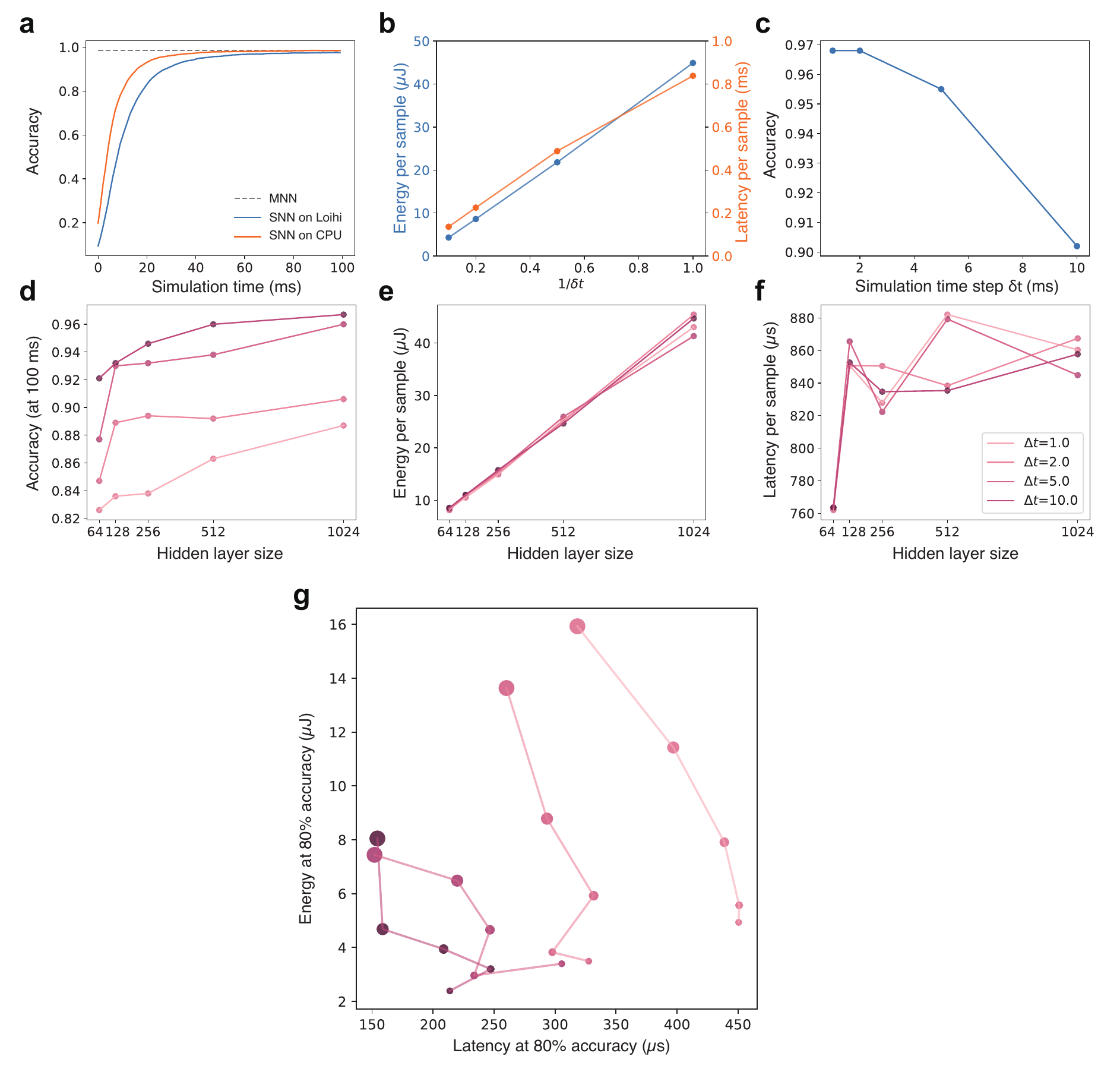}
    \caption{
    \textbf{Performance of SNN deployed on neuromorphic chip. } 
    \textbf{a}, Classification accuracy increases with simulation time steps and converges to the theoretical limit predicted by the MNN. Compared to single-precision floating-point simulation on CPU, the simulation on Loihi has a small amount of accuracy loss caused by weight quantization. 
    \textbf{b}, The classification accuracy over time of each model.
    \textbf{c}, The classification accuracy (at 100 ms of simulation time) for varying hidden layer size and $\Delta t$ used in loss for training.
    \textbf{d}, The average energy cost per sample (at 100 ms of simulation time) for varying hidden layer sizes and $\Delta t$.
    \textbf{e}, The average latency (at 100 ms of simulation time) for varying hidden layer size and $\Delta t$.
    \textbf{f}, Energy-latency diagram (at 80\% accuracy) revealing a trade-off between energy cost and latency. The dot size corresponds to hidden layer size and the color corresponds to the value of $\Delta t$ used during training. 
    } 
    \label{fig:loihi}
\end{figure}

We first check the consistency in prediction accuracy between the SNN implemented on Loihi and that implemented on CPU. As shown in Fig.~\ref{fig:loihi}\textbf{a}, the prediction accuracy of the SNN model deployed on Loihi increases with time and approaches to the theoretical limit predicted by the MNN (dashed line in Fig.~\ref{fig:loihi}\textbf{a}). Compared to the single-precision floating-point implementation on CPU, the Loihi implementation has a small amount of accuracy loss caused by the 9-bit weight quantization. Nonetheless, the result is impressive considering no fining tuning of free parameters are applied after training. %

On neuromorphic hardware, it is often desirable to use a coarser time step $\delta t$ to reduce energy consumption and delay. As shown in Fig.~\ref{fig:loihi}\textbf{b}, both energy consumption and delay (in wall time) are roughly inversely proportional to $\delta t$ for fixed simulation duration ($T=100$ ms). However, the accuracy on neuromorphic hardware deteriorates with coarser time discretization (Fig.~\ref{fig:loihi}\textbf{c}). This is due to that the model parameters are trained under the assumption of continuous LIF model rather than its discrete version. In practice, we find that this accuracy deterioration is negligible for $\delta t$ much smaller than the membrane potential time constant $1/L=20$ ms. To strike a balance between accuracy and energy consumption, we set $\delta t=1$ ms for the remaining results. Despite the present limitation, the general approach of moment embedding can be extended in the future to overcome this problem by deriving directly the moment mapping for the discrete SNN model rather than the continuous one.

As indicated by our general theory of SNC, there is a direct relationship between the uncertainty expressed by MNN and the required decision time, which in turn impacts the energy consumption. This suggests a potential way to optimize the trade-off between accuracy, energy cost, and delay on neuromorphic hardware. In the following, we investigate how the energy cost and delay depend on the hyperparameters of MNN during training. All models are trained over 30 epochs and the energy profiles are calculated through the first 1000 samples of the test dataset. %

Specifically, we analyze how the hidden layer size and the readout time window $\Delta t$ used in the loss function [equation (\ref{eq:loss_CE})] affect the model performance on Loihi. Moment neural networks with one hidden layer of different sizes (for $n=$ 64, 128, 256, 512, 1024) are trained under different values of $\Delta t$ in the moment loss function (for $\Delta t =$ 1, 2, 5, 10 ms). These models' performances as measured by accuracy, energy consumption, and delay are summarized in Fig.~\ref{fig:loihi}. %

We first look at the performance metrics after the accuracy fully converges at $t=100$ ms. As shown in Fig.~\ref{fig:loihi}\textbf{d}, a larger hidden layer size in general leads to better accuracy, but the improvement becomes less significant at larger sizes. The final accuracy is also improved by a larger readout time $\Delta t$ in the moment cross entropy loss used during the training. This is consistent with our theoretical prediction as a smaller $\Delta t$ prioritize accuracy at shorter time scales at the detriment of the accuracy at long time scales. As shown in Fig.~\ref{fig:loihi}\textbf{e}, the energy consumption (measured as the dynamic energy per sample) increases linearly with the hidden layer size and is largely unaffected by $\Delta t$ used during training. This is expected as energy cost is approximately proportional to the number of spikes and to the network size provided that the overall firing rate is fixed. As shown in Fig.~\ref{fig:loihi}\textbf{f}, the latency per sample does not show a clear trend across different network size or $\Delta t$ (with an exception at very small network size of $n=64$). This is because the primary factor determining latency (wall time) is the model simulation time which is fixed at $t=100$ ms.

As can be seen from results above, the performance as measured by accuracy, energy consumption, and delay have complex dependency on model hyperparameters, suggesting a potential way to optimize the trade-off between them. Specifically, we would like to know how we can achieve the best energy-delay product for any given level of accuracy. For this purpose, we identify the simulation time when each model reaches 80\% of accuracy and then record the corresponding energy consumption and delay. The results are best visualized using an energy-delay diagram as shown in Fig.~\ref{fig:loihi}\textbf{h}. Each curve represents the energy-delay characteristic profile for achieving a fixed level of accuracy as the hidden layer size varies. We find that a larger hidden layer size generally leads to less latency but at the cost of higher energy, whereas larger $\Delta t$ generally leads to better efficiency in both energy consumption and latency. Using the energy-delay diagram we can conveniently choose the appropriate network size to achieve an optimal trade-off between energy consumption and latency.

\section{Performance of the MNN on datasets other than MNIST}
\qi{
To demonstrate the general applicability and scalability of our method, we conduct extra experiments on the Fashion-MNIST and CIFAR-10 datasets with deeper networks. The Fashion-MNIST dataset consists of grayscale images of clothing items belonging to 10 categories, with 60,000 images for training and 10,000 images for validation. The CIFAR-10 dataset consists of natural images belonging to 10 categories, with 50,000 training images and 10,000 validation images.}

\qi{
We adopt a feedforward model configuration similar to the one used for the MNIST dataset as presented in the main text, and encode the pixel intensities of each image into Poisson spike trains. For the FashionMNIST dataset, our models consist of an input layer with 784 neurons, a hidden layer with 1000 neurons, and a linear readout with 10 units. For the CIFAR-10 dataset, the models consist of an input layer with 3072 neurons, three hidden layers (each with 1000 neurons), and a linear readout with 10 units. We train MNN models on these datasets with three different loss functions: the moment cross-entropy (MCE) loss with readout times \(\Delta t = 1\) or 10 ms, and the standard cross-entropy (CE) loss. As a reference for comparison, we also train conventional artificial neural networks (ANNs) with ReLU activation and identical network architectures (using CE loss). In both tasks, we employ AdamW with default hyperparameters (a learning rate of 0.001 and a weight decay of 0.01) to train these models for 30 epochs. Furthermore, we reconstruct spiking neural networks (SNNs) from the trained MNN models and evaluate the classification accuracy of the corresponding SNNs by running simulations for 100 ms with a time increment of \(\delta t = 1\) ms, averaged over 100 trials. The prediction accuracy of these models on the test sets is summarized in table ~\ref{tab:extra_experiments}. We find that for all tasks, the performance of the MNN is comparable to its ANN counterpart under the same network structure. The performance of the reconstructed SNN is reasonably close to the MNN, considering that no extra parameters are introduced both during and after training. 
}

\qi{We have not included larger datasets like CIFAR-100 or ImageNet as it would require an extension of the MNN to convolutional architectures. This extension is conceptually straightforward, as it would require replacing the matrix multiplication in equation (\ref{eq:sum_cov}) with a double-sided convolution. However, there is currently no available convolution kernel for covariance and developing efficient code infrastructure is a direction of future works.}

\begin{table}[]
\captionsetup{justification=centering}
\centering
\caption{Models' performance on MNIST, FashionMNIST and CIFAR10 datasets}
\label{tab:extra_experiments}
\begin{tabular}{ccccc}
\hline\hline
        \textbf{Dataset}   & \textbf{Model} & \textbf{Loss function} & \textbf{Accuracy (\%)} & \textbf{SNN's accuracy (\%)} \\ \hline
\multirow{4}{*}{MNIST} & \multirow{3}{*}{MNN} & MCE (\(\Delta t= 1\)) & 98.58 & 98.58 \\ 
                  &  & MCE (\(\Delta t = 10\)) & 98.68 & 98.71 \\
                  &  & CE & 98.63 & 98.68 \\ \cline{2-5} 
                  & ANN & CE & 98.82 & - \\ \hline
\multirow{4}{*}{FashionMNIST} & \multirow{3}{*}{MNN} & MCE (\(\Delta t= 1\)) & 86.74 & 86.66 \\ 
                  &  & MCE (\(\Delta t = 10\)) & 87.52 & 87.46 \\
                  &  & CE & 87.53 & 87.48 \\ \cline{2-5} 
                  & ANN & CE & 87.57 & - \\ \hline
\multirow{4}{*}{CIFAR10} & \multirow{3}{*}{MNN} & MCE (\(\Delta t = 1\)) & 53.67 &  
 51.23 \\ 
                  &  & MCE (\(\Delta t = 10\)) & 55.83 & 51.33 \\ 
                  &  & CE & 56.19 & 55.37 \\ \cline{2-5} 
                  & ANN & CE & 61.12 & - \\ \hline
\end{tabular}
\end{table}

\section{Quantitative comparison to previous methods for training SNN}

\qi{
In this section, We provide a quantitative comparison to previous methods for training SNN on the MNIST dataset under similar conditions as ours (table.~\ref{tab:model_comparison}). The following metrics are compared: accuracy for continuous-valued model (ANN and MNN) and the accuracy for SNN after convergence given sufficient time, inference time per sample, the total number of spikes per inference. The parameter settings including network size, membrane time constant, and firing threshold, and the techniques used are also commented. We find that while all methods summarized here can achieve similar performance in terms of the final accuracy, our method is superior in terms of inference time and the number of spikes (proportional to energy cost on neuromorphic hardware). Previously, best inference time and energy saving are achieved with post-training weight normalization. Here, the MNN provides an alternative approach with competitive performance without any post-training optimization or extra parameters during training.
}

\qi{
For meaningful comparison, we have only included results based on fully connected feedforward networks and exclude convolutional networks, as currently there is no available code infrastructure for convolution kernel for covariance computation. While generalizing the standard convolution kernel to second-order is conceptually straightforward, its efficient numerical implementation is better suited for a future study. For this reason, we have not included comparisons with CIFAR-10 as this dataset is most commonly solved using convolution networks~\citep{ding2021optimal,Yan2021,Hunsberger2015arxiv}. We also restrict our comparison to models using rate coding and exclude those using time coding or rank-order coding~\citep{Wunderlich2021sci_rep,Yan2021}. We have included comparisons to gradient-based learning methods, both conversion-based and direct, and excluded comparisons to methods using spike-time dependent plasticity. 
}

\qi{While we can’t directly compare the training results to that presented in ref.~\citep{10.3389/fnins.2017.00682}, as they consider exclusively convolutional neural networks, it is instructive to compare the general approach and the theories (which is independent of the specific model architecture). In brief, the authors in ref.\citep{10.3389/fnins.2017.00682} analyzed the relationship between the sample spike count and the input rate, and identified the source of error during ANN-to-SNN conversion (see Eq. 5 of their paper). They then showed how various techniques (specifically, weight normalization, using subtraction rather than hard reset in SNN, and using analog input instead of Poisson spikes) can be used to improve ANN-to-SNN conversion by eliminating this conversion error. The essential difference of this approach from ours is that these techniques are applied after training is done, whereas our model does not require any post-training optimization. This is because the MNN represents a \textit{tight} approximation to SNN in the first place. In other words, the conversion error is already minimized before training, if not completely eliminated. This is why we refer our approach as ‘reconstruction’ of SNN from MNN as opposed to ‘conversion’ of ANN to SNN. However, it is important to point out that our method does not exclude post-training optimization. In the future, both techniques can be applied in tandem to further boost performance in SNN.}

\qi{There are a number of benefits by considering probabilistic coding, both in terms of learning and inference. For learning, by considering the second-order moments of the spiking neurons, the immediate benefit is that we obtain the MNN as a tight approximation to the SNN, allowing us to reconstruct the SNN without any post-training optimization (as pointed out above). Another benefit is that by considering the stationary statistics, the MNN avoids the need to deal with temporal dynamics and thus does not require backpropagation-through-time. For inference, our approach can reduce the variance in the model prediction by manipulating correlated neural activity (in line with theories of neural population code), and in turn improve inference speed.}

\renewcommand{\arraystretch}{1.3}
\begin{table}[h] %
 \centering %
 \caption{Comparison of different methods for training spiking neural network on the MNIST dataset. IF indicates integrate-and-fire model without leak.}
 \label{tab:model_comparison} %
\begin{tabular}{@{} p{1.6cm}  p{2cm} p{1.8cm} p{1.8cm} p{1.8cm} p{1.3cm} p{1.3cm} p{3cm} @{}} %
\hline\hline 
  & Network
  structure & Accuracy
  \% (ANN/SNN) & Inference
  time (ms) & \#spikes
  & $\tau_{\rm m}$ (ms) & $V_{\rm th}$ (mV) & Techniques \\
\hline
\textbf{Conversion} &   &   &   &   &   &   &   \\
\hline
Hunsberger
  et al.~\citep{Hunsberger2015arxiv} & 784-500-200-10 & -/98.37 & n/a & n/a & 20
   & 20
   & Activation
  smoothing; Additive noise \\
Diehl ~~~~~~~~~~ et al.~\citep{Diehl2015} & 784-1200-1200-
  10 & 98.68/
  98.64 & 6$^\dagger$  %
   & 1146*
    & IF & 1
   & Weight
  normalization \\
  &   & 98.68/
  98.48 & 200* & 2525* & IF & 4
   & Manual
  adjustment of firing threshold \\
\hline
\textbf{Direct} &   &   &   &   &   &   &   \\
\hline
Lee ~~~~~~~~~~ et al.~\citep{Lee2016} & 784-800-10 & -/98.71 & 1000 & n/a & 20
   & adaptive & BPTT
  with error normalization;
  weight
  and threshold regularization \\
Wu ~~~~~~~~~~et al.~\citep{10.3389/fnins.2018.00331} & 784-800-10 & -/98.89 & 30 & n/a & 0.1
  ms & 1.5
   & STBP \\
Yan ~~~~~~~~~~
  et al.~\citep{Yan2021} & 784-256-64-10 & -/98.5 & 10 & 1148 & 1.11
  ms & 1.5
   & G-STBP;
  Sparsity regularization \\
\hline
\textbf{Reconstruction} &   &   &   &   &   &   &   \\
\hline
MNN (this study) & 784-1000-10 & 98.45/
98.43 & 6.6$^\dagger$ %
& 985 (300 in hidden layer) & 20
  ms & 20
   & Moment
  embedding \\
\hline \\
\multicolumn{8}{l}{* Estimated from Fig. 2 of their paper.}\\
\multicolumn{8}{l}{$\dagger$ At 99.6\% of maximum accuracy.}\\
\end{tabular}
\end{table}

\section{Comparison of spike count distributions between MNN and SNN}
\qi{
To further demonstrate that the MNN faithfully captures the distributions of spiking activity of the reconstructed SNNs, we compare the spike count distribution obtained from simulations of the SNN to that predicted by the MNN. To do so, we run simulations of the SNN to perform inference on the MNIST dataset, using the same model structure as presented in the main text, and record the spike count of hidden layer neurons over a 1000 ms time window across 500 trials. For this purpose, a time step of $\delta t = 0.01$ ms is used for simulating the SNN. 
}

\begin{figure}
    \centering
    \includegraphics[width=0.8\textwidth]{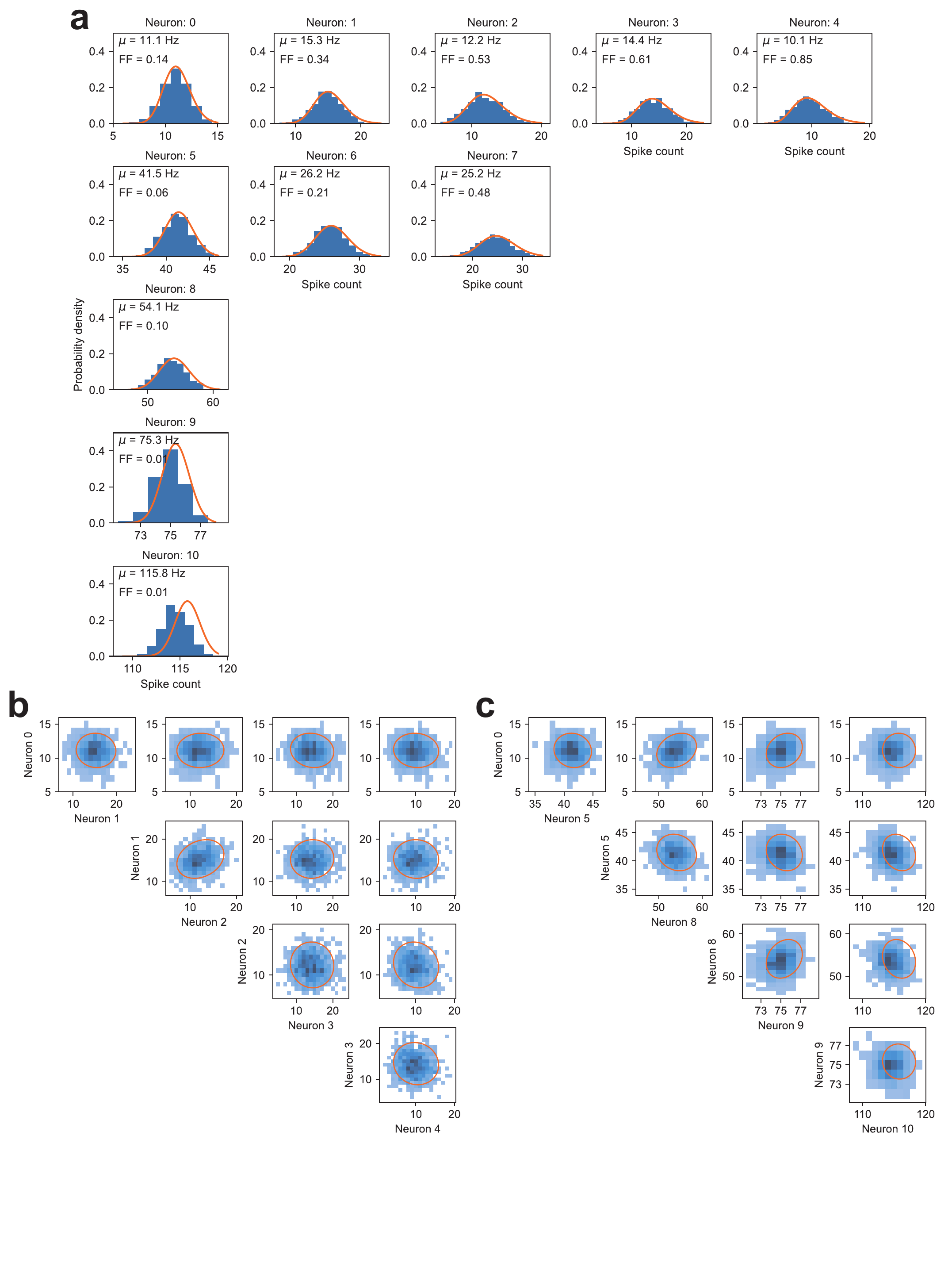}
    \caption{Comparing spike count distributions between MNN and SNN.     
    \textbf{a}, Spike count distribution of representative neurons in the hidden layer. Histograms corresponds to spike counts obtained from simulations of the SNN over 1000 ms time window across 500 trials. Orange curve shows gamma distributions with mean and variance calculated from the MNN. 
    \textbf{b}, Marginal joint probability density of the spike counts of pairs of neurons with similar mean firing rates (neuron 0, 1, 2, 3, 4).
    \textbf{c}, Marginal joint probability density of the spike counts of pairs of neurons with similar Fano factors (neuron 0, 5, 8, 9, 10). Heat map indicates the probability density obtained from simulations of the SNN; orange ellipse indicates covariance calculated by the MNN (scaled to 2 standard deviations).
    } 
    \label{fig:snn_dist}
\end{figure}

\qi{
Figure~\ref{fig:snn_dist}\textbf{a} shows the spike count distributions of 10 representative neurons (out of 1000 hidden neurons) for a randomly picked input image. The neurons shown cover a wide range of mean firing rate and Fano factors. Specifically, we first group the neurons according to their mean firing rates (five bins in the range of 0-150 Hz) and Fano factors (five bins in the range of 0-1), and then pick a random neuron from each group. The panels in Figure~\ref{fig:snn_dist}\textbf{a} are organized in a way that the mean firing rates increase from top to bottom whereas the Fano factors increase from left to right, with each row corresponding to neurons with similar mean firing rate and each column to neurons with similar Fano factors. Note that no neuron is found with both a high firing rate and a high Fano factor. We find that the spike count histograms as obtained from simulations of the SNN match well to the predictions by the MNN (orange curves, which show gamma distributions with mean and variance given by the MNN).
}

\qi{
To show the MNN also captures the joint distributions, we show the joint histograms between pairs of neurons and compare them to the prediction by the MNN. Figure~\ref{fig:snn_dist}\textbf{b} and Fig.~\ref{fig:snn_dist}\textbf{c} show the marginal joint probability density functions of the spike counts of pairs of neurons with similar mean firing rates (neuron 0, 1, 2, 3, 4) and similar Fano factors (neuron 0, 5, 8, 9, 10), respectively. The heat maps indicate the histograms obtained from simulations of the SNN and the orange ellipses indicate the covariance calculated by the MNN (scaled to 2 units of standard deviation).
}

\end{document}